\documentclass[10pt,journal,compsoc]{IEEEtran}

\ifCLASSOPTIONcompsoc
\usepackage[nocompress]{cite}
\else

\usepackage{cite}
\fi

\ifCLASSINFOpdf
\usepackage[pdftex]{graphicx}

\graphicspath{{../pdf/}{../jpeg/}}

\else

\fi

\usepackage[cmex10]{amsmath}

\usepackage{algorithm}
\usepackage{fancybox}
\usepackage[noend]{algpseudocode}
\usepackage{caption}  
\usepackage{subcaption}
\begin{document}

\title{Egocentric Meets Top-view}

\author{Shervin~Ardeshir, and Ali~Borji,~\IEEEmembership{Member,~IEEE,} 
\IEEEcompsocitemizethanks{\IEEEcompsocthanksitem S. Ardeshir and Ali Borji are with the Department
of Computer Science, University of Central Florida, Orlando,
FL, 32816.\protect\\
web page: see crcv.ucf.edu/~borji and http://www.shervin-ardeshir.com \protect\\ 
This work updates and extends our previous ECCV 2016 paper\cite{ardeshir2016ego2top}.
}
}

\markboth{}%
{Shell \MakeLowercase{\textit{et al.}}: Bare Demo of IEEEtran.cls for Computer Society Journals}

\IEEEtitleabstractindextext{%
\begin{abstract}
Thanks to the availability and increasing popularity of wearable devices such as GoPro cameras, smart phones, and glasses, we have now access to a plethora of videos captured from the first person perspective. Surveillance cameras and Unmanned Aerial Vehicles (UAVs) also offer tremendous amounts of video data recorded from top and oblique view points. Egocentric and surveillance vision have been studied extensively but separately in the computer vision community. The relationship between these two domains, however, remains unexplored. In this study, we make the first attempt in this direction by addressing two basic yet challenging questions. First, having a set of egocentric videos and a top-view video, does the top-view video contain all or some of the egocentric viewers? In other words, have these videos been shot in the same environment at the same time? Second, if so, how can we identify the egocentric viewers in the top-view video? These problems can become even more challenging when videos are not temporally aligned. We model each view (egocentric or top) using a graph, and compute the assignment and time-delays in an iterative-alternative fashion using spectral graph matching and time delay estimation. Such an approach handles the temporal misalignment between the egocentric videos and the top-view video. We evaluate our method in terms of ranking and assigning egocentric viewers to identities present in the top-view camera over a dataset of 50 top-view and 188 egocentric videos captured under different conditions. Results demonstrate the effectiveness of the proposed  iterative-alternative graph-based approach. We believe that our work is an important first step towards bridging egocentric and surveillance domains and is useful for many future works and applications. 
\end{abstract}

}

\maketitle

\IEEEdisplaynontitleabstractindextext

\IEEEpeerreviewmaketitle

\IEEEraisesectionheading{\section{Introduction}\label{sec:introduction}}

\IEEEPARstart{O}{n} one hand, wearable devices such as GoPro cameras, smart phones, and glasses have recently provided us with a large amount of video data from the first person point of view. Analysis of these videos has become an interesting and rapidly-growing research area in computer vision, from detecting and recognizing daily actions (e.g.,~\cite{egoActionFathi,egoDailyAction}) to localizing the field of view of an egocentric viewer (e.g., \cite{egoFOVLocalization}). The human-centric nature of egocentric vision offers the opportunity to study computer vision from our perspective which is the first person point of view.

On the other hand, surveillance cameras and unmanned aerial vehicles capture a lot of visual information about daily activities and events taking place in different locations over long periods of time. Surveillance and generally top-view vision has a long history in the computer vision research, from human detection and re-identification (e.g., \cite{DPM1,DPM2,reidReimannian}) to object tracking (e.g.,~\cite{ZamirECCV12}).

These two types of visual data, capturing drastically different view-points, provide complementary sources of information. If combined correctly, together they can provide rich analytical power. A thorough understanding of this relationship can open the door to adapting the extensive amount of research done on third person vision to the new area of egocentric vision. Further, establishing such a relationship can have several important applications. For instance, videos of athletes equipped with body-worn cameras alongside with videos captured by static top-view cameras can offer additional insights for sport analysis which might not be available from each individual source. As another example, finding the person behind an egocentric camera in a surveillance network could be useful for law enforcement given the increasing use of wearable devices by police officers. Furthermore, fusing these two types of information, egocentric and surveillance, can result in a better 3D reconstruction of an environment. Another use case would be helping visually-impaired people, equipped with egocentric cameras, in tasks such as navigation or obstacle avoidance (e.g.,~\cite{pradeep2010wearable}).

\begin{figure}[t]
	\centering
	\includegraphics[width=1\linewidth]{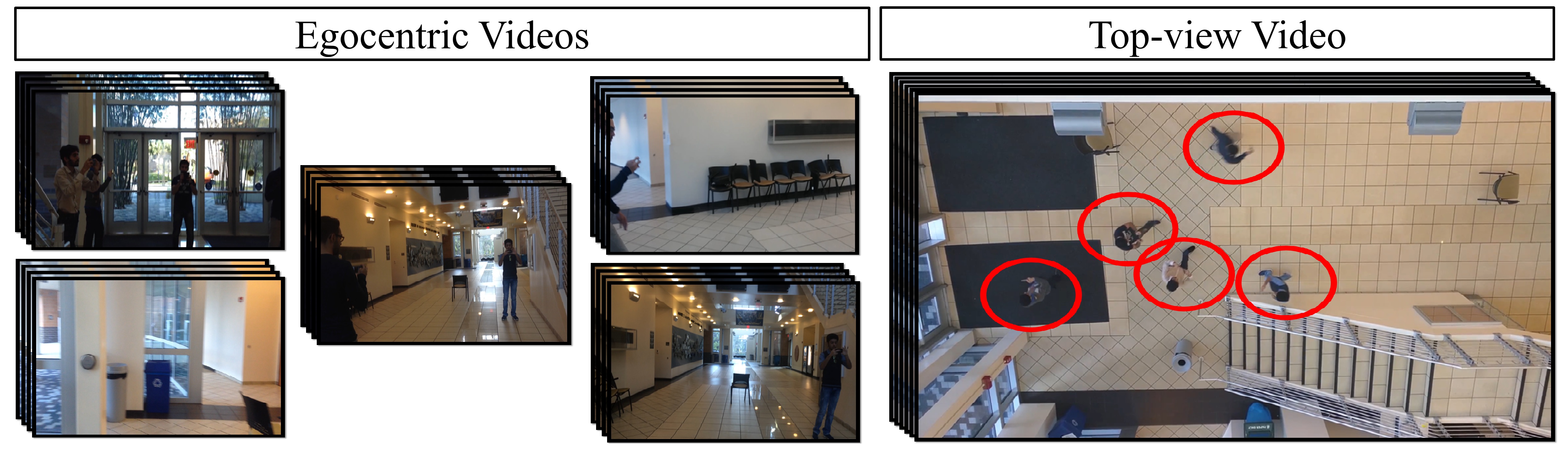}
	\caption{Left) a set of 5 egocentric videos. Right) a top-view video capturing the scene and possibly the egocentric viewers. The viewers are highlighted using red circles in the top-view video. We aim to answer the two following questions: 1) Does this set of egocentric videos belong to the viewers visible in the top-view video? 2) Assuming they do, which viewer is capturing which egocentric video?}	
	\label{fig:Example}
\end{figure}

The first principal step towards relating the egocentric and top-view vision, is to establish correspondences between them. Efficiently matching the content between egocentric and top-view cameras is necessary for additional mutual analysis of both contents. 
To take the first step in this direction, we consider a specific scenario which is localizing and identifying people recording the egocentric videos in a top-view reference camera, as illustrated in Figure \ref{fig:Example}. 

We ask the two following questions. Given a set of egocentric videos and a top-view surveillance video: 1) Does this set of egocentric videos belong to the viewers visible in the top-view camera? and 2) If yes, then which viewer is capturing which egocentric video? To answer these questions, we need to compare a set of egocentric videos to a set of viewers visible in a single top-view video. To find a matching, in our solution, each set is represented by a graph and the two graphs are compared using a spectral graph matching technique \cite{spectralMatching}. In the egocentric graph, each egocentric video is a node. In the top-view graph, each node corresponds to a visible viewer. In general, this problem can be very challenging due to the nature of egocentric cameras. The camera-holder is not visible in his own egocentric video which leave us with no cues about his visual appearance. 

In order to evaluate our method, we use the same dataset by Ardeshir and Borji~\cite{ardeshir2016ego2top,ardeshiregocentric}. It contains several test sets. In each set, multiple people, hereafter referred to as ego-centric \textit{viewers}, are walking around while recording videos. Simultaneously, a top-view camera is recording the entire area including all or some of the egocentric viewers and possibly other intruders (See Figure \ref{fig:Example}). In what follows, we mention some challenges concerning this problem and sketch the layout of our approach.

\begin{figure*}[t]
	\centering
	\includegraphics[width=1\linewidth]{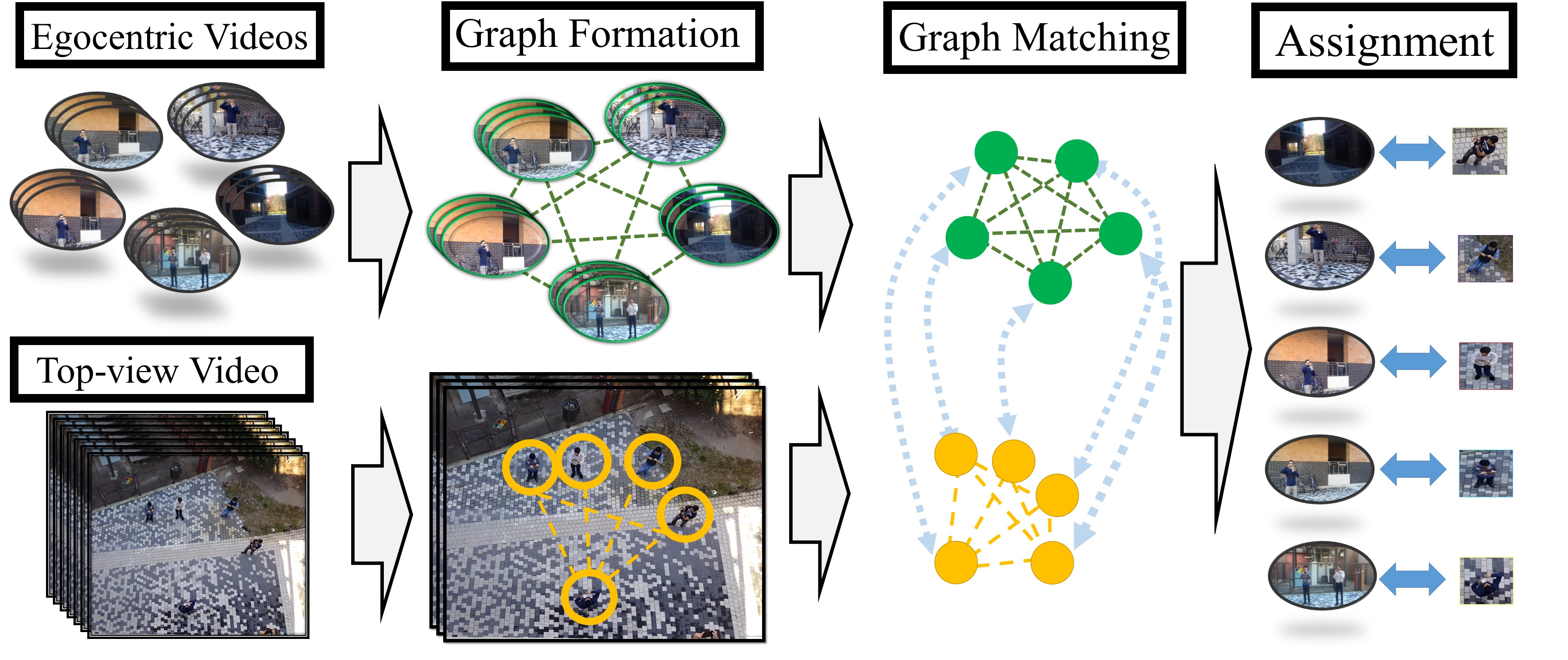}
	\caption{The input to our framework is a set of egocentric videos (in this case 5 videos), and one top-view video. The goal is defined as assigning the egocentric videos to the people recording them. A graph is formed on the set of egocentric videos (each node being one of the egocentric videos), and another graph is formed on the top-view video (each node representing one of the targets present in the video). Using spectral graph matching, a soft assignment is found between the two graphs, and using a soft-to-hard assignment, each egocentric video is assigned to one of the viewers in the top-view video. This assignment is our answer to the second question in Figure \ref{fig:Example}.}
	\label{fig:teaser}
\end{figure*} 

In order to have an understanding of the behavior of each individual in the top-view video, we use a multiple object tracking method \cite{theWayTheyMove} to extract the viewer's trajectory in the top-view video. Note that an egocentric video captures a person's field of view rather than his spatial location. Therefore, the content of a viewer's egocentric video, a 2D scene, corresponds to the content of the viewer's field of view in the top-view camera. For the sake of brevity, we refer to a viewer's top-view field of view as Top-FOV in what follows. Since trajectories computed by multiple object tracking do not provide us with the orientation of the egocentric cameras in the top-view video, we assume that for the most part humans tend to look straight ahead (i.e., front-looking head and torso) and therefore shoot videos from the world in front of them. This is usually the case when viewers wear the camera on their body (Please see Figure \ref{fig:inFOVs}). Having an estimate of a viewer's orientation and Top-FOV, we then encode the changes in his Top-FOV over time and use it as a descriptor. We show that this feature correlates with the change in the global visual content (or Gist) of the scene observed in his corresponding egocentric video. 

We also define pairwise features to capture the relationship between a pair of egocentric videos, and similarly the relationship between a pair of viewers in the top-view camera. Intuitively, if an egocentric viewer observes a certain scene and another egocentric viewer comes across the same scene some time later, this could hint as a relationship between the two cameras. If we match a top-view viewer to one of the two egocentric videos, we are likely to be able to find the other viewer using the mentioned relationship. As we experimentally show, this pairwise relationship significantly improves our assignment accuracy. This assignment will lead us to define a score measuring the similarity between the two graphs. Our experiments demonstrate that the graph matching score could be used for verifying if the top-view video is in fact capturing the egocentric viewers (See the diagram shown in Figure \ref{fig:sceneMatching_teaser}). 

The rest of this paper is organized as follows. In Section \ref{sec:related_work}, we mention related works to our study. In Section \ref{sec:framework}, we describe details of our framework. Section \ref{sec:results} presents our experimental results followed by discussions and conclusions in Section \ref{sec:conclusion}.
\begin{figure*}[t]
	\begin{center}
		\includegraphics[width=.8\textwidth]{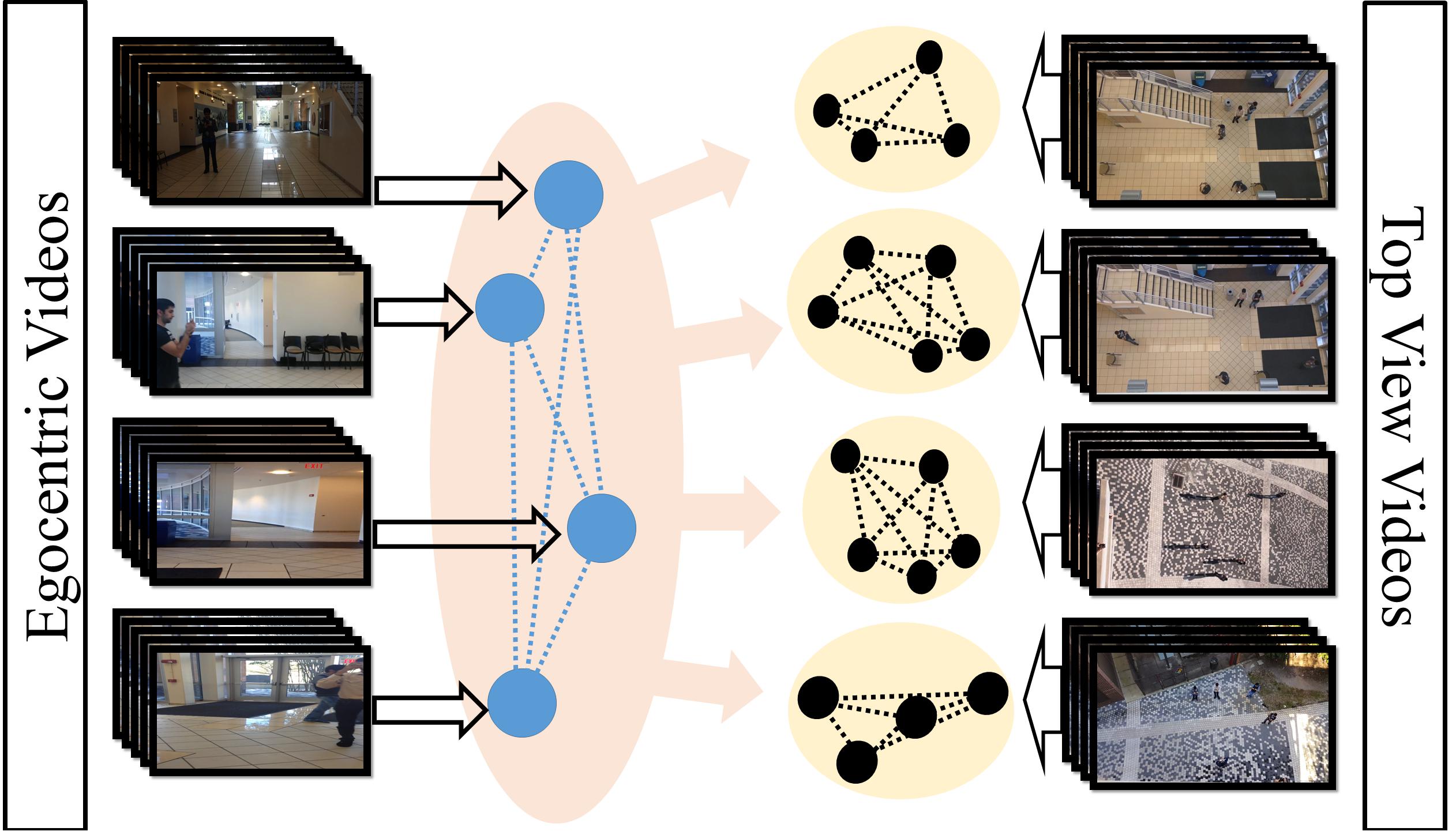}			
		\caption{Adapting our method for evaluating top-view videos. We form a graph on the set of egocentric videos and compare this graph to other graphs built on different top-view videos. The top-view videos are ranked based on how similar their graph is to the egocentric graph. The performance of this ranking helps us answer our first question.}
		\label{fig:sceneMatching}			
	\end{center}
\end{figure*}

\section{Related Work}
\label{sec:related_work}
Visual analysis of egocentric videos has recently became a hot research topic in computer vision~\cite{egoKanade,egoEvolutionSurvey}, from recognizing daily activities \cite{egoDailyAction,egoActionFathi} to object detection \cite{egoObjectDetection}, video summarization \cite{egoVideoSummarization}, and predicting gaze behavior~\cite{egoli2013learning,egoPolatsekNovelty,Borji2014look}. In the following, we review some previous work related to ours spanning \textit{Relating static and egocentric},~\textit{Social interactions among egocentric viewers}, and~\textit{Person identification and localization}.\\

\noindent\textbf{Relating Static and Egocentric Cameras:} Some studies have addressed relationships between moving and static cameras. 
Interesting works reported in \cite{egoMobileFixedObjectDetection,egoMobileFixedMasterSlave} have explored the relationship between mobile and static cameras for the purpose of improving object detection accuracy. \cite{egoExo} fuses information from egocentric and exocentric vision (third-person static cameras in the environment) with laser depth range data to improve depth perception and 3D reconstruction. Park {\em et al.} \cite{egoPredictingGaze} predict gaze behavior in social scenes using first-person and third-person cameras. Soran {\em et al.},~\cite{soran2014action} have addressed action recognition in presence of an egocentric video and multiple static videos.\\

\noindent\textbf{Social Interactions among Egocentric Viewers:}
To explore the relationship among multiple egocentric viewers, \cite{egoWisdomOfTheCrowd} combines several egocentric videos to achieve a more complete video with less quality degradation by estimating the importance of different scene regions and incorporating the consensus among several egocentric videos. Fathi {\em et al.},~\cite{egoSocialInteractions} detect and recognize the type of social interactions such as dialogue, monologue, and discussion by detecting human faces and estimating their body and head orientations. Yonetani {\em et al.} \cite{yonetani2015ego} correlate the head motion of an egocentric observer with the humans present in other egocentric videos to perform self-search. \cite{egoMultiTaskClustering} proposes a multi-task clustering framework, which searches for coherent clusters of daily actions using the notion that people tend to perform similar actions in certain environments such as workplace or kitchen. \cite{egoYouDoILearn} proposes a framework that discovers static and movable objects used by a set of egocentric users. Recent work in \cite{lin2015co} identifies the person who draws the most attention in a set of egocentric viewers, given a set of time-synchronized egocentric videos interacting with each other. \\

\noindent\textbf{Person Identification and Localization:} Perhaps, the most similar computer vision task to ours is person re-identification \cite{reidCPS,reidReimannian,reidSDALF,Assari2016reidentification}. The objective here is to find and identify people across multiple cameras. In other words, who is each person present in one static camera, in another overlapping or non-overlapping static camera? However, the main cue in human re-identification is visual appearance of humans, which is absent in egocentric videos. 
Tasks such as human identification and localization in egocentric cameras have been studied in the past. \cite{egoHeadMotion} uses the head motion of an egocentric viewer as a biometric signature for determine which videos have been captured by the same person. In \cite{egoSurfing}, authors identify egocentric observers in other egocentric videos, using their head motion. 
The work of \cite{egoFOVLocalization} localizes the field of view of an egocentric camera by matching it against a reference dataset of videos or images (such as Google street view). 
Landmarks and map symbols have been used in~\cite{egoWhereAmI} to perform self localization on the map. The study reported in \cite{chakraborty2016person} addresses the problem of person re-identification in a surveillance network of wearable devices, and \cite{zhengidentifying} performs re-identification on time-synchronized wearable cameras.

\section{Framework}
\label{sec:framework}
The block diagram in Figure \ref{fig:teaser} illustrates different steps of our approach. 
\textbf{First}, each view (ego-centric or top-down) is represented by a graph which defines the relationship among the viewers present in the scene. These two graphs may not have the same number of nodes for two reasons: a) some of the egocentric videos might not be available, b) some individuals, present in the top-view video, might not be capturing videos. Each graph consists of a set of nodes where each node represents a viewer (egocentric or top-view), and each edge represents a pairwise relationship between two viewers.

We represent each viewer in the top-view by describing his expected Top-FOV, and in egocentric view by the visual content of his video over time. These descriptions are encoded in the graph nodes. We also define pairwise relationships between pairs of viewers, which are encoded as the edge features of the graph (i.e., how two viewers' visual experience relate to each other).

\textbf{Second}, we use spectral graph matching to compute a score measuring the similarity between the two graphs, alongside with an assignment from the nodes of the egocentric graph to the nodes of the top-view graph. Since the videos are not necessarily time-synchronized, it is important to take the relative time-delays between the videos into account. Therefore, we propose an iterative method, which simultaneously estimates the assignments and the relative time-delays between the egocentric viewers and the top-view video. We try two different iterative-alternative algorithms, analyze the pros and cons of each, and evaluate their performance on our dataset.

Our experiments show that the graph matching score can be used as a measure of similarity between the egocentric graph and the top-view graph. As a result, it can be used as a measure for verifying whether a set of egocentric videos are recorded by the viewers visible in the top-view video. Therefore, it allows us to evaluate the capability of our method in terms of answering our first question. In addition, the assignment obtained by the graph matching suggests an answer to our second question. We organize this section by first describing the graph formation process for each of the views, and then describing the details of the matching procedure.


\subsection{Graph Representation} Each view, egocentric or top-view, is described using a single graph. 
The set of egocentric videos is represented using a graph in which each node represents one of the egocentric videos, and an edge captures the pairwise relationship between the content of the two videos. 

In the top-view graph, each node represents the expected visual experience of a viewer being tracked (in the top-view video), and an edge captures the pairwise relationship between the two visual experiences over time. \textit{Visual experience} refers to what a viewer is expected to observe during the course of his recording seen from the top view camera.\\

\noindent \textbf{3.1.1 \ \ Modeling the Top-View Graph:} In order to model the visual experience of a viewer in the top-view camera, knowledge about his spatial location (i.e., trajectory) throughout the video is needed. We employ the multiple object tracking method presented in \cite{theWayTheyMove} to extract a set of trajectories, each corresponding to one of the viewers in the scene. Similar to \cite{theWayTheyMove}, we use annotated bounding boxes, and provide their centers as an input to the multiple object tracker. Our tracking results here are nearly perfect due to several reasons: the high quality of videos, high video frame rate, and lack of challenges such as occlusion in the top-view videos.  

\begin{figure*}[t]
	\begin{center}
		\begin{subfigure}{0.33\textwidth}
			\includegraphics[width=\textwidth]{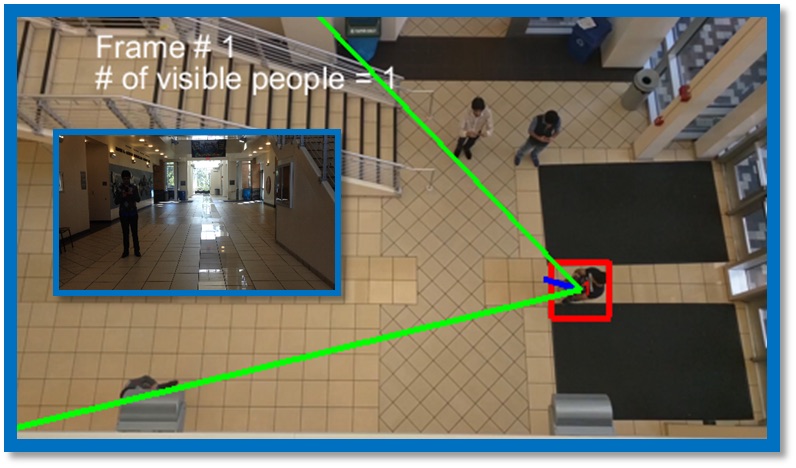}
			\caption{}		
		\end{subfigure}\hfill	
		\begin{subfigure}{0.33\textwidth}
			\includegraphics[width=\textwidth]{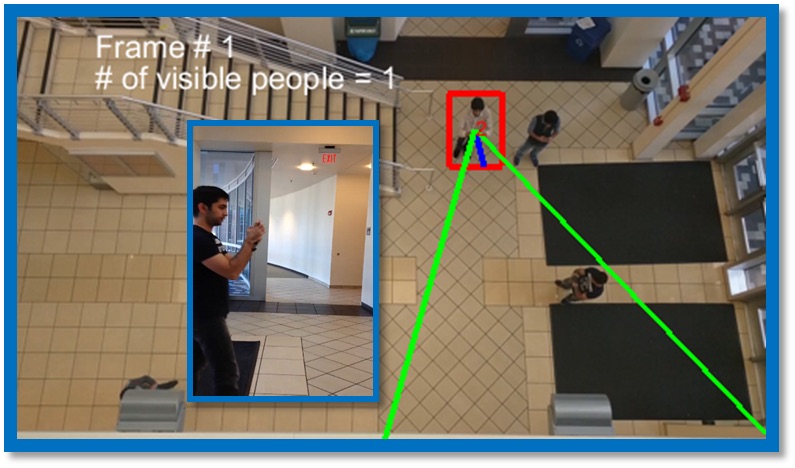}	
			\caption{}	
		\end{subfigure}\hfill	
		\begin{subfigure}{0.33\textwidth}
			\includegraphics[width=\textwidth]{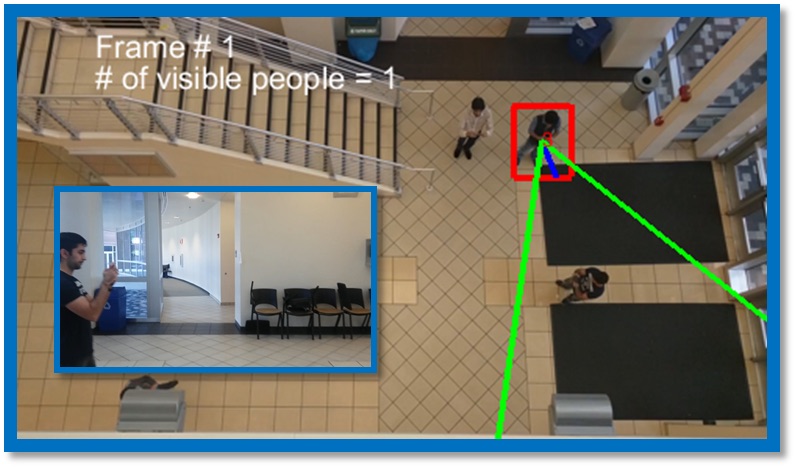}	
			\caption{}	
		\end{subfigure}\hfill									
		\caption{Expected field of view for three different viewers in the top-view video alongside with their corresponding egocentric frames. The short dark blue line shows the estimated orientation of the camera. The Top-FOV shown in (b) and (c) have a high overlap, therefore we expect their egocentric videos to have relatively similar visual content compared to the pairs (a,b) or (a,c) at this specific time.}
		\label{fig:inFOVs}			
	\end{center}	
\end{figure*}

Each node represents one of the individuals being tracked. Employing the general assumption that people often tend to look straight ahead, we use a person's speed vector as the direction of his camera at time t (denoted as $\theta_t$). Further, assuming a fixed angle ($\theta_d$), we expect the content of the person's egocentric video to be consistent with the content included in a 2D cone formed by the two rays emanating from the viewer's location and with angles $\theta - \theta_d$ and $\theta + \theta_d$. Figure \ref{fig:inFOVs} illustrates the expected Top-FOV for three different individuals present in a frame. In our experiments, we set $\theta_d$ to 30 degrees. In theory, angle $\theta_d$ can be estimated more accurately by knowing intrinsic camera parameters such as focal length and sensor size of the corresponding egocentric camera. However, since we do not know the corresponding egocentric camera, we set it to a default value.

Top-FOVs are not directly comparable to viewers' egocentric views. The area in the Top-FOV in a top-view video mostly contains the ground floor which is not what an ego-centric viewer usually observes in front of him. However, what can be used to compare the two views is the relative change in the Top-FOV of a viewer over time. This change should correlate with the change in the content of the egocentric video. Intuitively, if a viewer is looking straight ahead while walking on a straight line, his Top-FOV is not going to have drastic changes. Therefore, we expect the viewer's egocentric view to have a stable visual content.\\

\noindent\textbf{Node Features:} We extract two unary features for each node, one captures the changes in the content covered by his FOV, and the other is the number of visible people in the content of the Top-FOV. 

To encode the relative change in the visual content of viewer $i$ visible in the top-view camera, we form the $T \times T$ matrix ($T$ denotes the number of frames in the top-view video) $U^{IOU}_{i}$ whose elements $U^{IOU}_{i}(f_p,f_q)$ indicate the IOU (intersection over union) of the Top-FOV of person $i$ in frames $f_p$ and $f_q$. For example, if the viewer's Top-FOV in frame 10 has high overlap with his FOV in frame 30 (thus $U^{IOU}_{i}(10,30)$ has a high value), we expect to see a high visual similarity between frames 10 and 30 in the egocentric video. Two examples of such features are illustrated in the middle column of Figure \ref{fig:FOVvsGIST} (a). 

\begin{figure*}
   	\centering
   	\includegraphics[width=1\linewidth]{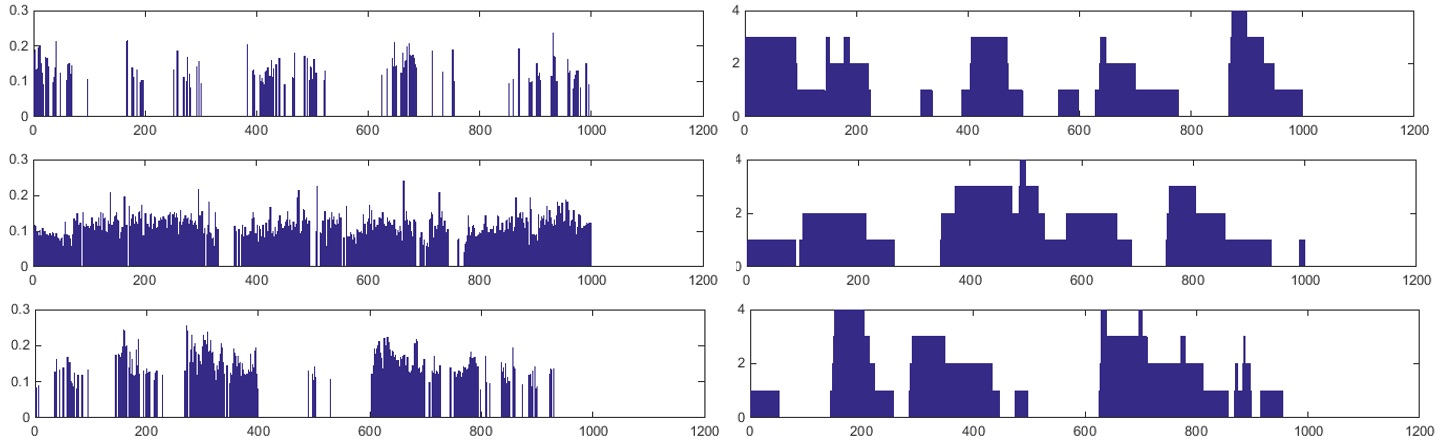}
   	\caption{Examples of 1D features capturing the number of humans in different frames of the videos. Left shows the summation of the detection scores at every frame for their corresponding egocentric videos. Right panel shows the number of visible people in three different viewer's Top-FOV over time. The x axis encodes the frame number in which the number of humans was measured. Both vectors are normalized and then compared to each other. The similarity between the two patterns shows the discriminative power of this feature, especially if the video is long enough (as it has 1000 frames in this case). However, our experiments show that in most cases where human detection results are not that confident or the video length is too short, this feature by itself does not results in a high assignment accuracy. }
   	\label{fig:features_1D}
   \end{figure*}

Having the Top-FOV of viewer $i$ estimated, we then count the number of people within his Top-FOV at each time frame and store it in a ${1 \times T}$ vector $U^{n}_{i}$. To compute the number of visible people, we count the number of annotated bounding boxes within his Top-FOV. Figure \ref{fig:inFOVs} illustrates three viewers who have one human in their Top-FOV. A few examples of this feature are visualized in the top row of Figure \ref{fig:features_1D}.\\

\noindent\textbf{Edge Features:} Pairwise features are designed to capture the relationship among two different individuals. In the top-view videos, similar to the unary matrix $U^{IOU}_{i}$, we can form a $T \times T$ matrix $B^{IOU}_{ij}$ to describe the relationship between a pair of viewers (viewers/nodes $i$ and $j$), in which $B^{IOU}_{ij}(f_p,f_q)$ is defined as the intersection over union of the Top-FOVs of person $i$ in frame $f_p$ and person $j$ in frame $f_q$. Intuitively, if there is a high similarity between the Top-FOVs of person $i$ in frame $f_p$ and person $j$ in frame $f_q$, we would expect the $f_q$th frame of viewer $j$'s egocentric video to be similar to the $f_p$th frame of viewer $i$'s egocentric video. Two examples of such features are illustrated in the middle column of Figure \ref{fig:FOVvsGIST} (b). \\

\noindent \textbf{3.1.2 \ \ Modeling the Egocentric Graph:} As in the top-view graph, we also construct a graph on the set of egocentric videos. Each node of this graph represents one of the egocentric videos. Edges between the nodes capture the relationship between a pair of egocentric videos.\\

\noindent\textbf{Node Features:} Similar to the top-view graph, each node is represented using two features. First, we capture how the overall visual experience is evolving. We compute pairwise similarity between GIST features \cite{GIST} of all video frames (for one viewer) and store the pairwise similarities in a $T_{E_i} \times T_{E_i}$ matrix $U^{GIST}_{E_i}$, in which the element $U^{GIST}_{E_i}(f_p,f_q)$ is the GIST similarity between frame $f_p$ and $f_q$ of egocentric video $i$, and $T_{E_i}$ is the number of frames in the $i$th egocentric video. Two examples of such features are illustrated in the left column of Figure \ref{fig:FOVvsGIST} (a). The GIST similarity is a function of the euclidean distance of the GIST feature vectors. 
\begin{equation}
U^{GIST}_{E_i}(f_1,f_2)=e^{-\gamma|g^{E_i}_{f_p}-g^{E_i}_{f_q}|}.
\end{equation}
In which $g^{E_i}_{f_p}$ and $g^{E_i}_{f_q}$ are the GIST descriptors of frame $f_p$ and $f_q$ of egocentric video $i$, and $\gamma$ is a constant which we empirically set to $0.5$.

The second feature is a time series counting the number of visible people in each frame. In order to have an estimate of the number of people, we run a pre-trained human detector using deformable part model DPM~\cite{DPM2} on each egocentric video frame. In order to make sure that our method is not including humans in far distances (which are not likely to be present in the top-view camera), we exclude bounding boxes whose sizes are smaller than a certain threshold (determined considering an average human height of 1.7m and distance of the diameter of the area being covered in the top view video.).
Each of the remaining bounding boxes, has a detection score which is rescaled into the interval [0 1]. The rescaled score has the notion of the probability of that bounding box containing a person. Scores of all detections in a frame are added and used as a count of people in that frame. Therefore, similar to the top-view feature, we can represent the node $E_i$ of egocentric video $i$ with a $1 \times T_{E_i}$ vector $U^{n}_{E_i}$.
A few examples of this feature are visualized in the bottom row of Figure \ref{fig:features_1D}.\\

\noindent\textbf{Edge Features:} To capture the pairwise relationship between egocentric cameras $i$ (containing $T_{E_i}$ frames) and $j$ (containing $T_{E_j}$ frames), we extract GIST features from all of the frames of both videos and form a $T_{E_i} \times T_{E_j}$ matrix $B^{GIST}_{ij}$ in which $B^{GIST}_{ij}(f_p,f_q)$ represents the GIST similarity between frame $f_p$ of video $i$ and frame $f_q$ of video $j$. 
\begin{equation}
B^{GIST}_{ij}(f_p,f_q)=e^{-\gamma|g^{E_i}_{f_p}-g^{E_j}_{f_q}|}.
\end{equation}

Two examples of such features are illustrated in the left column of Figure \ref{fig:FOVvsGIST} (b). 

\subsection{Graph Matching}
Our goal in this section is to find a binary assignment matrix $\mathbf{X}_{N^e \times N^t}$, in which $N^e$ is the number of egocentric videos and $N^t$ is the number of people in the top-view video. $\mathbf{X}(i,j)$ equal to 1 means that egocentric video $i$ has been matched to viewer $j$ in the top-view video. To capture the similarities between the elements of the two graphs, we define the affinity matrix $A_{N^eN^t \times N^eN^t}$ in which $a_{ik,jl}$ is the affinity of edge $ij$ in the egocentric graph with edge $kl$ in the top-view graph. Reshaping matrix $\mathbf{X}$ as a vector $\mathbf{x}_{N^eN^t \times 1} \in \{0,1\}^{N^eN^t}$, the assignment problem could be defined as maximizing the following objective function:
\begin{equation}
\underset{x}{\operatorname{argmax}} \ \mathbf{x}^T\mathbf{A}\mathbf{x}.
\end{equation} 
We compute $a_{ik,jl}$ based on the similarity between the feature descriptor of edge $ij$ in the egocentric graph $B^{GIST}_{ij}$ and the feature descriptor for edge $kl$ in the top-view graph $B^{IOU}_{kl}$. Once the affinity matrix is known we can measure the probability of each of the nodes in the first graph being matched to each of the nodes in the second graph. This probabilistic assignment is commonly known as soft-assignment.

\noindent\textbf{Soft Assignment} We employ the spectral graph matching method introduced in \cite{spectralMatching} to compute a soft assignment between the set of egocentric viewers and top-view viewers. In \cite{spectralMatching}, assuming that the affinity matrix is an empirical estimation of the pairwise assignment probability, and the assignment probabilities are statistically independent, $\mathbf{A}$ is represented using its rank one estimation which is computed by: 
\begin{equation}
\underset{\mathbf{p}}{\operatorname{argmin}} \ |\mathbf{A}-\mathbf{p}\mathbf{p}^T|.
\end{equation}
In fact, the rank one estimation of $\mathbf{A}$ is no different than its leading eigenvector. Therefore, $\mathbf{p}$ can be computed either using eigen decompositon, or estimated iteratively using power iteration. Considering vector $\mathbf{p}$ as the assignment probablities, we can reshape $\mathbf{p}_{N^eN^t \times 1}$ into a $N^e \times N^t$ soft assignment matrix $\mathbf{P}$, for which  $\mathbf{P}(i,j)$ represents the probability of matching egocentric viewer $i$ to viewer $j$ in the top-view video after row normalization. \\
  
\noindent\textbf{Hard Assignment} Any soft to hard assignment method can be used to convert the soft assignment result (generated by spectral matching) to the hard binary assignment between the nodes of the graphs. We used the well-known Munkres (also known as Hungarian) algorithm \cite{kuhn1955hungarian} to obtain the final binary assignment. 
\\
In the following, we first describe our previous method introduced in \cite{ardeshir2016ego2top} which solely solves the viewer assignment (section \ref{sec:ECCV_method}). We then describe our new two iterative algorithms in section \ref{sec:PAMI_method}, which aims to simultaneously estimate the time-delays and find the best assignments.

\subsection{Solving Viewer Assignment}
\label{sec:ECCV_method}
As described in the previous section, each of the nodes and edge features is a 2D matrix. $B^{GIST}_{ij}$ is a $T_{E_i} \times T_{E_j}$ matrix, $T_{E_i}$ and $T_{E_j}$ are the number of frames in egocentric videos $i$ and $j$, respectively. $B^{IOU}_{kl}$ is a $T_{t} \times T_{t}$ matrix and $T_{t}$ denotes the number of frames in the top-view video. Note that $B^{GIST}_{ij}$ and $B^{IOU}_{kl}$ are not directly comparable as the two matrices are not of the same size (the videos do not necessarily have the same length). Also, the absolute time in the videos do not correspond to each other as the videos are not time-synchronized. In fact, the relationship between viewers $i$ and $j$ in the 100th frame of the top-view video does not correspond to frame number 100 of the egocentric videos. Due to this, we expect to see a correlation between the GIST similarity of frame $100+d_i$ of egocentric video $i$ and frame $100+d_j$ of egocentric video $j$, and the intersection over union of in Top-FOVs of viewers $k$ and $l$ in frame 100. $d_i$ and $d_j$ are the time delays of egocentric videos $i$ and $j$ with respect to the top-view video.

\begin{figure*}[t]
   	\begin{center}
   		\begin{subfigure}[t]{0.48\textwidth}
   			\includegraphics[width=1\linewidth]{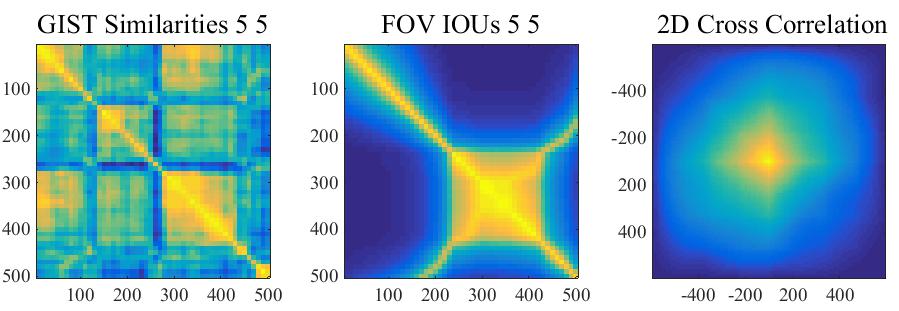}	
   		\end{subfigure}\hspace{0.15 in} 
   		\begin{subfigure}[t]{0.48\textwidth}
   			\includegraphics[width=1\linewidth]{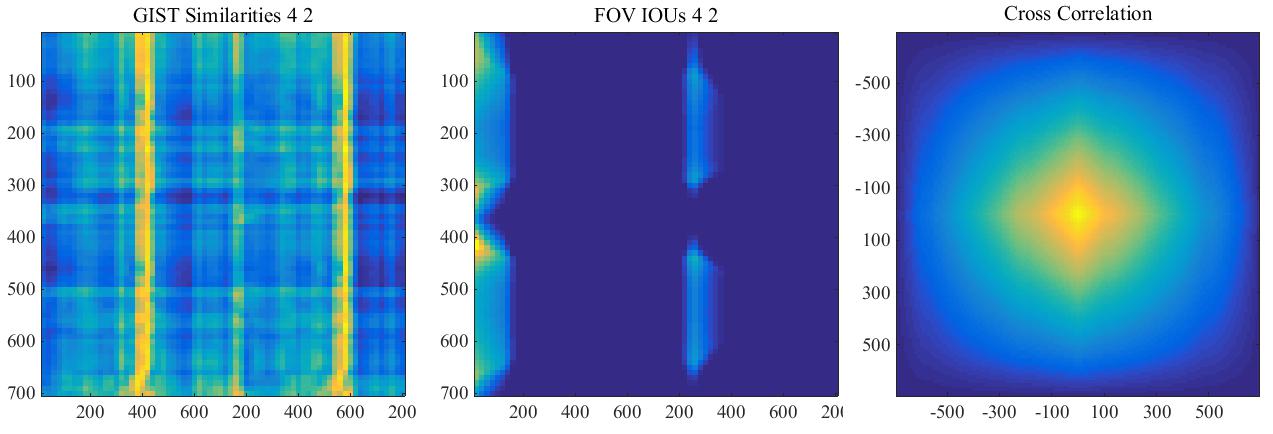}	
   		\end{subfigure}
   		
   		\begin{subfigure}[t]{0.48\textwidth}
   			\includegraphics[width=1\linewidth]{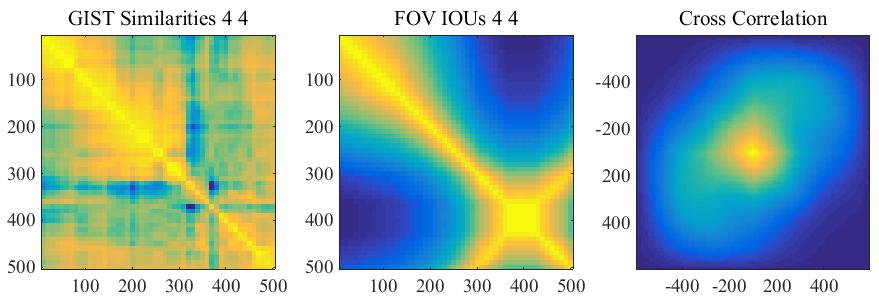}
   			\caption{}
   		\end{subfigure}\hspace{0.15 in}
   		\begin{subfigure}[t]{0.48\textwidth}
   			\includegraphics[width=1\linewidth]{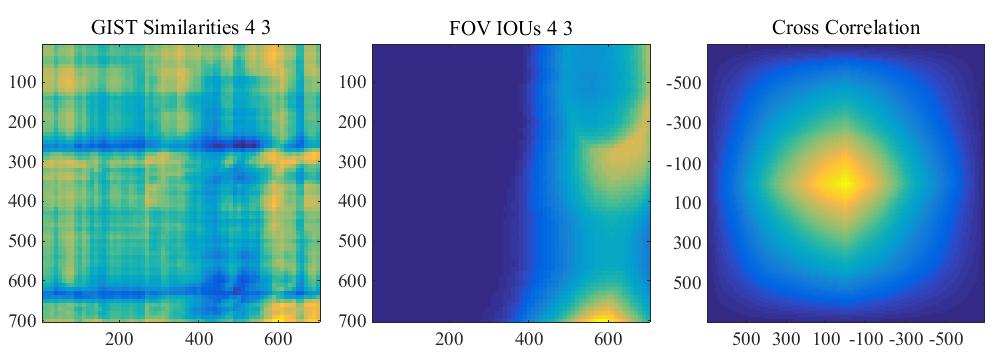}
   			\caption{}	
   		\end{subfigure}		
   		\caption{(a) shows two different examples of the 2D features extracted from the \textbf{nodes} of the graphs for which the values are color-coded. Left column shows the 2D matrices extracted from the pairwise similarities of the GIST feature descriptors $U^{GIST}$, middle shows the 2D matrices computed by intersection over union of the FOV in the top-view camera $U^{IOU}$, and the rightmost column shows the result of the 2D cross correlation between the two. (b) shows the same concept, but between two \textbf{edges}. Again, the leftmost figure shows the pairwise similarity between GIST descriptors of one egocentric camera to another $B^{GIST}$. Middle, shows the pairwise intersection over union of the FOVs of the pair of viewers $B^{IOU}$, and the rightmost column illustrates their 2D cross correlation. The similarities between the GIST and FOV matrices in fact capture the affinity of two nodes/edges in the two graphs.}
   		\label{fig:FOVvsGIST}										
   	\end{center}
   \end{figure*}

In \cite{ardeshir2016ego2top}, the affinity between two edges is defined as the following:

\begin{equation}
A_{ikjl}=\text{max} (B^{GIST}_{ij} \ast B^{IOU}_{kl}).
\label{eq:binary_affinity}
\end{equation}

where $\ast$ denotes cross correlation.  
For the elements of $A$ for which $i=j$ and $k=l$, the affinity captures the compatibility of node $i$ in the egocentric graph, to node $k$ in the top-view graph. The compatibility between the two nodes is computed using 2D cross correlation between $U^{IOU}_{k}$ and $U^{GIST}_{E_i}$ and 1D cross correlation between $U^{n}_{k}$ and $U^{n}_{E_i}$. The overall compatibility of the two nodes is a weighted linear combination of the two:
\begin{equation}
\label{eq:node_similarity}
A_{ikik}=\alpha \text{max} (U^{GIST}_{E_i} \ast U^{IOU}_{k}) + (1-\alpha) \text{max}(U^{n}_{E_i} \ast U^{n}_{k}),
\end{equation}
where $\alpha$ is a constant between 0 and 1 specifying the contribution of each term. In our experiments, we set $\alpha$ to 0.9. Figure \ref{fig:FOVvsGIST} illustrates the features extracted from some of the nodes and edges in the two graphs.
Where maximum of cross correlation occurs is interpreted as the best offset(delay) which makes the two matrices the most similar. The time delay problem is handled properly by assuming each cross-correlation is maximized on an offset equal to the time-delays of its corresponding egocentric videos. This assumption might not always hold as it does not enforce consistency among the assumed time-delays. We will address this issue using the approaches described in the next section.   

\subsection{Joint Optimization Over Assignment and Time-delays}
\label{sec:PAMI_method}
\begin{figure*}
	\begin{center}
		\begin{subfigure}[t]{0.9\textwidth}
			\includegraphics[width=1\linewidth]{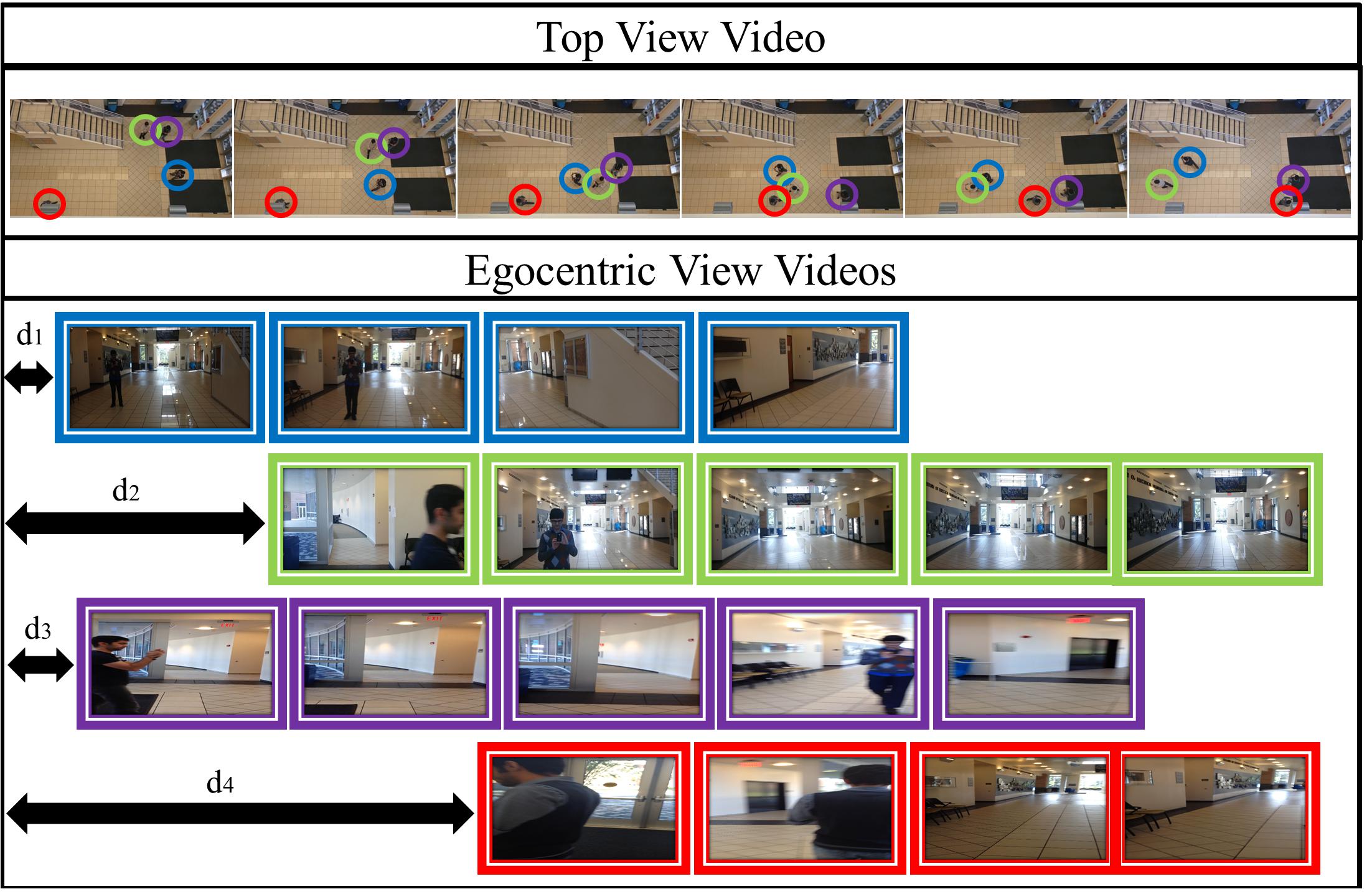}
			\label{fig:example1}
		\end{subfigure}
		\begin{subfigure}[t]{0.9\textwidth}
			\includegraphics[width=1\linewidth]{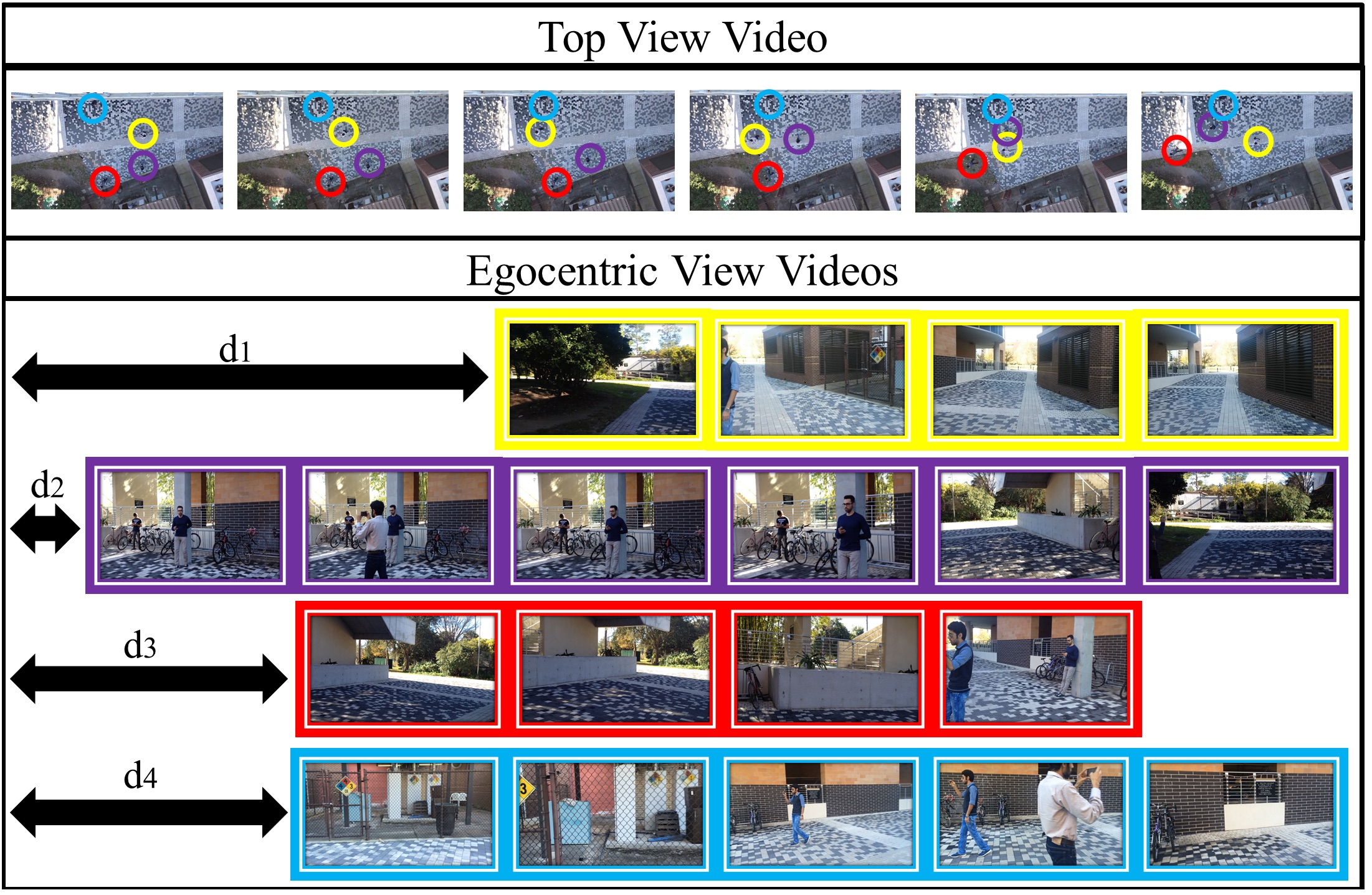}
			\label{fig:example2}
		\end{subfigure}
		\caption{Two test case examples: A few sample frames of the top-view video, and each of the egocentric videos are visualized alongside with their time-delays with respect to the top-view video time-reference. The color-coding denotes the identities of the viewers.}
		\label{fig:example_of_test_cases}
	\end{center}
\end{figure*}

The shortcoming of the similarity definition in \cite{ardeshir2016ego2top} is that it does not enforce any sort of consistency among the time-delays assigned to different egocentric videos. In fact, the problem of viewer assignment, and time-delays of the egocentric videos are interconnected. On one hand, we need to have an estimation of the time-delays, to be able to correctly measure the node-to-node/edge-to-edge similarities of the corresponding nodes/edges. On the other hand, we need to know the correct assignment to be able to estimate the time-delay between two videos. Theoretically, if we assume the top-view video as a reference of absolute time (as shown in figure \ref{fig:example_of_test_cases}), each cross-correlation maximization is suggesting one(for nodes), or two(for edges) egocentric time delays with respect to the top-view video's absolute time. As an example, if the edge between egocentric videos $i$ and $j$ has its cross correlation with its corresponding top-view edge maximized at $d_i'$ and $d_j'$, that suggests those values for the time-delay of egocentric videos $i$, and $j$. Therefore, if the cross-correlation of edge $ik$ being maximized in the first dimension at $d_i''$(which $d_i'' \ne d_i'$), we are assuming egocentric video $i$ is starting at two different absolute times, which is self-contradictory. Therefore, the framework needs to enforce consistency among the time-delays of all the egocentric videos, suggesting a unique time-delay for each individual egocentric video. As a result, we define the objective to jointly optimize the time-delays and the assignment. Intuitively, putting constraints on the time-delays, will put constraints on the solution space, as some of the solutions using \cite{ardeshir2016ego2top} might implicitly assign invalid (inconsistent) time-delays to the egocentric videos. 
 
Having $n$ egocentric videos, we can represent their unknown time delays, using a $1 \times n$ vector $\mathbf{t}_d$. Taking the time delay into account the objective will have the following form:

\begin{equation}
	\underset{\mathbf{x},\mathbf{t}_d}{\operatorname{argmax}} \ \mathbf{x}^T\mathbf{A}({\mathbf{t}_d})\mathbf{x}.
\end{equation} 

This brings us back to the chicken and egg nature of the problem, which suggests an iterative-alternative approach. Initializing the time delays, estimating the assignments, and refining the time-delays based on that. Intuitively, we should seek the optimum assignment, in addition to a time delay for each egocentric video. $\mathbf{A}(\mathbf{t}_d)$ is the affinity matrix, assuming the $i_{th}$ egocentric video has time delay $\mathbf{t}_d(i)$. Changing $\mathbf{t}_d$ will alter the elements of the affinity matrix, and $\mathbf{x}$ will decide which elements of the affinity matrix should contribute to the graph matching score.
We employed two different methods for solving this objective. First, we suggest a faster algorithm which first seeks an optimal time delay vector, and then proceeds to the assignment problem. The second algorithm is an iterative-alternative method which goes back and forth between the assignment and time-delay estimation. 
\\

\noindent \textbf{Spectral Optimization:} In the first approach, we find an optimum $\mathbf{A}$ resulting from the optimal time delays for the egocentric videos, and then solve the assignment using the obtained affinity matrix. In other word we assume:
\begin{equation}
\underset{\mathbf{x},\mathbf{t}_d}{\operatorname{argmax}} \ \mathbf{x}^T\mathbf{A}({\mathbf{t}_d})\mathbf{x}=\underset{\mathbf{x}}{\operatorname{argmax}} \ \mathbf{x}^T \ \mathbf{A}(\mathbf{t}_d^*) \ \mathbf{x}.
\end{equation}  
In order to find the optimum $\mathbf{t}_d$, we use the intuition behind the concept of leading eigenvalue of the affinity matrix. In spectral graph theory, leading eigenvalue captures the strength of it's most dominant cluster. In other words, the larger the leading eigenvalue is, the stronger the main cluster becomes. Our graph matching method is based on the assumption that the affinity matrix is well estimated by its rank one approximation using its leading eigenvector. Therefore, the better the leading eigenvector represent the affinity matrix, the more confident our spectral graph matching will be. As a result, the best affinity matrix corresponds to the most dense main cluster, and therefore the largest leading eigenvalue. According to this intuition, we can find $\mathbf{t}_d^*$ using:
\begin{equation}
\mathbf{t}_d^*=\underset{\mathbf{t}_d}{\operatorname{argmax}} \ \lambda_{\mathbf{A}(\mathbf{t}_d)}.
\end{equation}  
For solving the objective function above, we initialize the time delays and iteratively refine them using a local search in the $n$ dimensional space of the time-delay vector. The details are explained in Algorithm 1. Effectively, first we evaluate neighboring time delay vectors by analyzing their corresponding affinity matrices. Having a $n$ dimensional time-delay vector, we compute its neighboring time-delay vectors, by changing one of its elements (time-delay of one of the egocentric videos) by a single unit (which we empirically set to 0.1 sec). For each neighbor, we compute the resulting affinity matrix and its leading eigenvalue. We pick the neighboring time delay vector with the maximum leading eigenvalue in the affinity matrix and effectively maximize $\frac{\partial \lambda_{\mathbf{A}_{t_d}}}{\partial t_d}$, and update time delays and the affinity matrix. The algorithm keeps iterating until one of the convergence criteria are met. We define the convergence criteria as either reaching a local maximum leading eigenvalue, or reaching the maximum number of iterations. Once the criteria are met, soft and hard assignments are computed using the computed optimum affinity matrix. We explore the effect of the two different initializations in terms of assignment and ranking and compare it with \cite{ardeshir2016ego2top}.
\\

\begin{algorithm}
	\caption{Spectral Optimization}\label{euclid}
	\textbf{Input} = $\mathbf{G},\mathbf{G'}, itr_{max}, \epsilon$ \\
	\textbf{Output} = $\mathbf{x},\mathbf{t}_d$	\\
	
	\begin{algorithmic}[1]
		\Procedure{Spectral Solver}{}
		\State Initialize $\mathbf{t}_{d}$
		\State compute the affinity matrix $\mathbf{A}_{t_d}$
		\While {$\frac{\partial \lambda_{\mathbf{A}_{t_d}}}{\partial t_d} > \epsilon$ and $itr<itr_{max}$} 
		\State{$\mathbf{t}_d=\mathbf{t}_d+\frac{\partial \lambda_{\mathbf{A}_{t_d}}}{\partial \mathbf{t}_d}$}	
		\State update $\mathbf{A}_{t_d}$
		\EndWhile 
		\State compute soft and hard assignment $\mathbf{p}_{t_d}$ and $\mathbf{x}_{t_d}$ 
		\State \textbf{return} $\textbf{t}_d$ and $S(\mathbf{t}_d)$.
		\EndProcedure
	\end{algorithmic}
\end{algorithm}

\noindent \textbf{Matching Score Based Optimization:} In our second approach, we attempt to find the optimal values for $\mathbf{t}_d$ and $\mathbf{x}$ simultaneously using an iterative-alternative approach. First, we initialize $\mathbf{t}_d$, which leads to a constant affinity matrix. Second, we compute the assignments using spectral graph matching. The assignment is then used for further refining the time delays. In other words, we observe how the graph matching score changes, using different neighboring time delay vectors and pick the best direction for the growth of the graph matching score (which is essentially $\frac{\partial S(\mathbf{t}_d)}{\partial \mathbf{t}_d}$). We go back and forth between the time-delays and assignments until our termination criteria is met. Similar to algorithm 1, the termination criteria is defined as reaching a local maximum or maximum number of iterations. The details of this approach are explained in Algorithm 2. Our experiments show that this method can have a more favorable accuracy compared to the first approach (Algorithm 1), with the cost of more computational complexity as each iteration consist of additional steps of computing the assignment vector $\mathbf{x}$. The performance of this algorithm will be compared to the first approach in the Experimental Results section.
\\

\begin{algorithm}
	\caption{Matching Score Based Optimization}\label{euclid}
	\textbf{Input} = $\mathbf{G},\mathbf{G'}, itr_{max}, \epsilon$ \\
	\textbf{Output} = $\mathbf{x},\mathbf{t}_d$	\\
	
	\begin{algorithmic}[1]
		\Procedure{Matching Score Based Solver}{}
		\State Initialize $\mathbf{t}_{d}$
		\State compute the affinity matrix $\mathbf{A}_{t_d}$, soft assignment $\mathbf{p}_{t_d}$, hard assignment $\mathbf{x}_{t_d}$ and graph matching score $S(t_d) = \textbf{x}_{t_d}^T\textbf{A}_{t_d}{\textbf{x}}_{t_d}$
		\While {$\frac{\partial S(t_d)}{\partial t_d} > \epsilon$ and $itr<itr_{max}$} 
		\State{$\mathbf{t}_d=\mathbf{t}_d+\frac{\partial S(\mathbf{t}_d)}{\partial \mathbf{t}_d}$}	
		\State update $\mathbf{A}_{t_d}$, $\mathbf{p}_{t_d}$, $\mathbf{x}_{t_d}$, and $S(\mathbf{t}_d)$
		\EndWhile 
		\State \textbf{return} $\textbf{t}_d$ and $S(\mathbf{t}_d)$.
		\EndProcedure
	\end{algorithmic}
\end{algorithm}

\noindent\textbf{Initializing time-delays:} Since we locally search for the best objective, the initialization plays a significant role in the final results. Two different initialization methods are considered. First, we initialize the vector $\mathbf{t}_d$ with a vector of zeros, assuming the videos are time-synchronized. Second, we empirically estimate the time-delays by computing the median of all the values suggested by the cross-correlations. As explained in \cite{ardeshir2016ego2top}, each cross-correlation maximization suggests a time-delay for each of the egocentric cameras, therefore, each of the $N^e$ node/edge involving node $i$, will have $N^t$ suggested time delays (once cross-correlated with them). For each cross correlation maximization (equation \ref{eq:binary_affinity}) two expectations are likely to happen: a) random time-delay values suggested by incorrect corresponding nodes/edges, or b) consistent time-delay values suggested by correct correspondences. Therefore, we initialize the time delay of node $i$ as the median of all the suggested values for that specific node. For instance, time delay of egocentric video $\mathbf{E}_i$ is initialized as the following:
\begin{equation}
\mathbf{t^*}_{d_i}=\widetilde{\mathbf{T}}_{d_i}
\end{equation}
where ${T}_{d_i}$ is the set of all implicitly suggested time-delays implicitly by the elements of the two graphs:
\begin{equation}
\mathbf{T}_{d_i}= \{\underset{\mathbf{t}_{d_i}}{\operatorname{argmax}} \ \  B^{GIST}_{ij} \ast B^{IOU}_{kl} | \forall \ j\leq N^e,k,l\leq N^t\}.
\end{equation}
We evaluate the effect of this initialization by comparing it to the results of initializing $\mathbf{t}_d$ as a vector of zeros.

%
%
%

\section{Experimental Results}
\label{sec:results}
In this section, we mention details of our experimental setup, collected data, evaluation measures as well as some baseline methods.
\subsection{Dataset}
We collected a dataset containing 50 test cases of videos shot in different indoor and outdoor environments. Each test case contains one top-view video and several egocentric videos captured by the people visible in the top-view camera. Two test case examples are shown in Figure \ref{fig:example_of_test_cases}. Depending on the included subset of egocentric cameras, we can generate up to 2,862 instances of our assignment problem (will be explained in more detail in Section 4.2.4). Overall, our dataset contains more than 225,000 frames. Number of people visible in the top-view cameras varies from 3 to 10, number of egocentric cameras varies from 1 to 6, and the ratio of number of available egocentric cameras to the number of visible people in the top-view camera varies from  0.16 to 1. Lengths of the videos vary from 320 frames (10.6 seconds) up to 3132 frames (110 seconds).

\subsection{Evaluation}
We evaluate our method in terms of answering the two questions asked in the Introduction section. First, given a top-view video and a set of egocentric videos, can we verify if the top-view video is capturing the egocentric viewers? We analyze the capability of our method in answering this question in Section 4.2.1. 

Second, knowing that a top-view video contains the viewers recording a set of egocentric videos, can we determine which viewer has recorded which egocentric video? We answered this question in Sections 4.2.2 and 4.2.3.
\subsubsection{Ranking Top-view Videos:} 
\label{sec:sceneRanking}
\begin{figure}
	\begin{center}
	\includegraphics[width=0.53\textwidth]{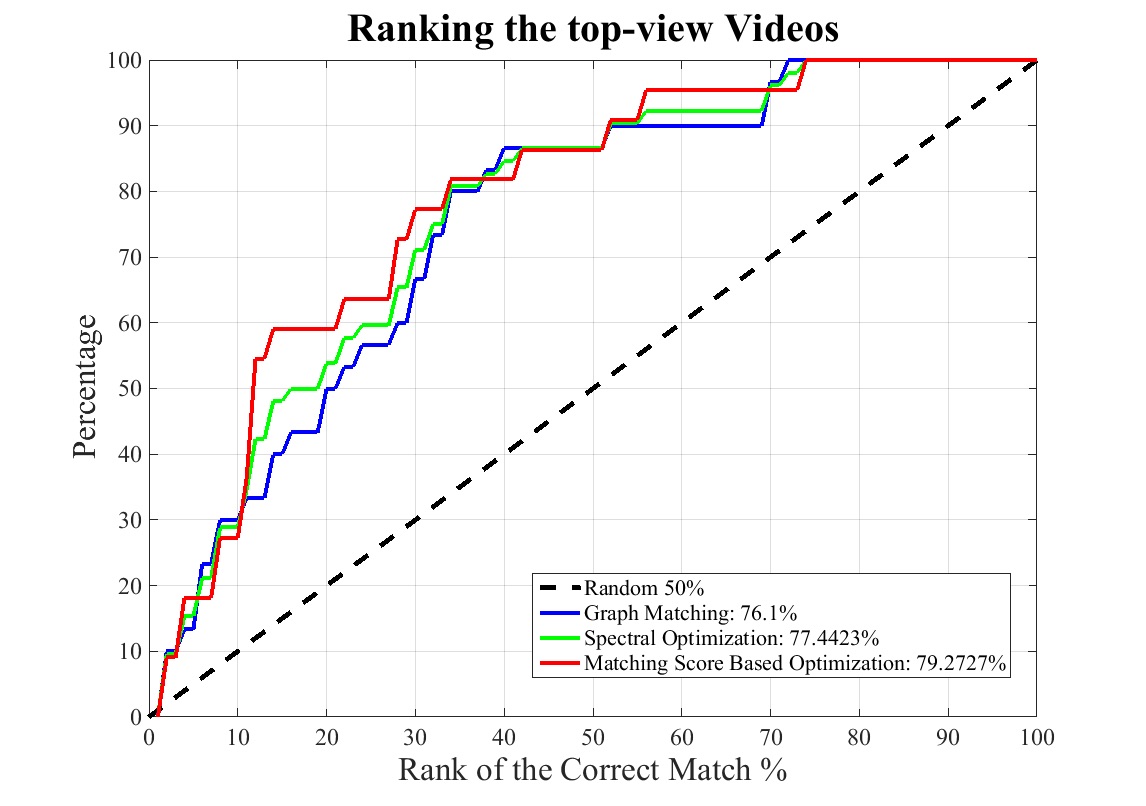}
	\caption{The cumulative matching curve demonstrates the performance of the proposed spectral, and matching based optimization methods (red and green), and compares them with the baseline graph matching method introduced in \cite{ardeshir2016ego2top}. It shows that our proposed algorithms outperform the baselines by more than 1.3\% and 3.1\%.}
	\label{fig:sceneMatching_results_PAMI}	
	\end{center}	
\end{figure}

We design an experiment to evaluate if our graph matching score is a good measure for the similarity between the set of egocentric videos and a top-view video. Having a set of egocentric videos from the same test case (recorded in the same environment), and 50 different top-view videos (from different test cases), we compare the similarity of each of the top-view graphs to the egocentric graph. After computing the hard assignment for each top view video(resulting in the assignment vector $x$), the score $x^TAx$ is associated to that top-view video. This score is effectively the summation of all similarities between the corresponding nodes and edges of the two graphs. All the top-view videos are evaluated and ranked using this score. The ranking accuracy is computed by measuring the rank of the ground truth top-view video, and computing the cumulative matching curves shown in Figure \ref{fig:sceneMatching_results_PAMI}. The blue curve shows the ranking accuracy when we apply the baseline method of \cite{ardeshir2016ego2top}, where time-delay consistency is not enforced. The green and red curve show the ranking accuracy of our proposed algorithms, spectral optimization and matching score based optimization respectively. The dashed black line shows the accuracy of randomly ranking the top-view videos. It can be observed that all the curves outperform the random ranking. This shows that our graph matching score is a meaningful measure for estimating the similarity between the egocentric videos and the set of viewers visible in the top-view video. In addition, the green and red curve outperforming the blue curve, indicates the effectiveness of our time-delay consistency enforcement. Also, the red curve giving us the best results, shows that our second algorithm outperforms the spectral level. Please note that both of the proposed methods were initialized using the medians of the suggested values as described in the initialization section. In general, this experiment answers the first question. Indeed, graph matching score can be used as a cue for narrowing down the search space among the top-view videos, for finding the one corresponding to our set of the egocentric cameras.

\subsubsection{Viewer Ranking Accuracy:}
We evaluate our soft assignment results, in terms of ranking capability. In other words, we can look at our soft assignment as a measure to sort the viewers in the top-view video based on their assignment probability to an egocentric video. Computing the ranks of the correct matches, we can plot the cumulative matching curves to illustrate their performance. 

We evaluate the performance of our proposed methods, each with two different initializations, and compare their performance with four baselines in Figure \ref{fig:all_results} (a). First, random ranking (dashed black line), in which for each egocentric video we randomly rank the viewers present in the top-view video. Second, sorting the top-view viewers based on the similarities of their 1D unary features to the 1D unary features of each egocentric camera (i.e., number of visible humans illustrated by the blue curve). Third, sorting the top-view viewers based on their 2D unary features (GIST vs. FOV, shown by the green curve). Note that here, we are ignoring the pairwise relationships (edges) in the graphs (the blue and green curves). The cyan curve illustrates the accuracy of the method used in \cite{ardeshir2016ego2top}, and the magenta and red curve shows the performance of our spectral based and graph matching score based methods. Sold curves are the outcome of median initialization, while the dashed curves are resulting from zero initialization. It can be observed that correctly initializing the time-delays has a significant impact on the performance.

\subsubsection{Assignment Accuracy:} 
\begin{figure*}[t]
	\begin{center}
		\begin{subfigure}[t]{0.49\textwidth}
			\includegraphics[width=\textwidth]{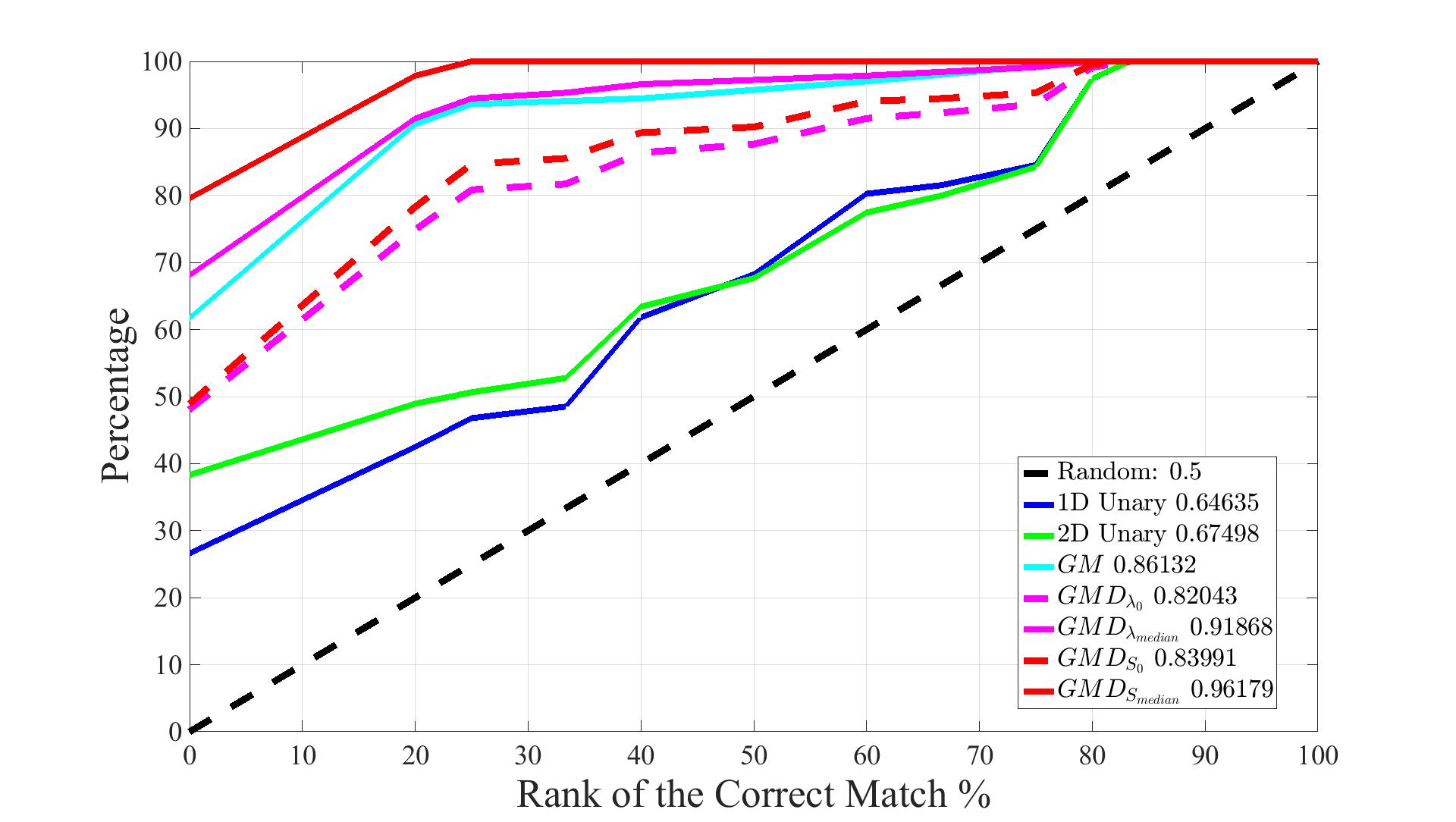}
			\caption{}
			\label{fig:fullSet_cmc}			
		\end{subfigure}
		\begin{subfigure}[t]{0.51\textwidth}
			\includegraphics[width=\textwidth]{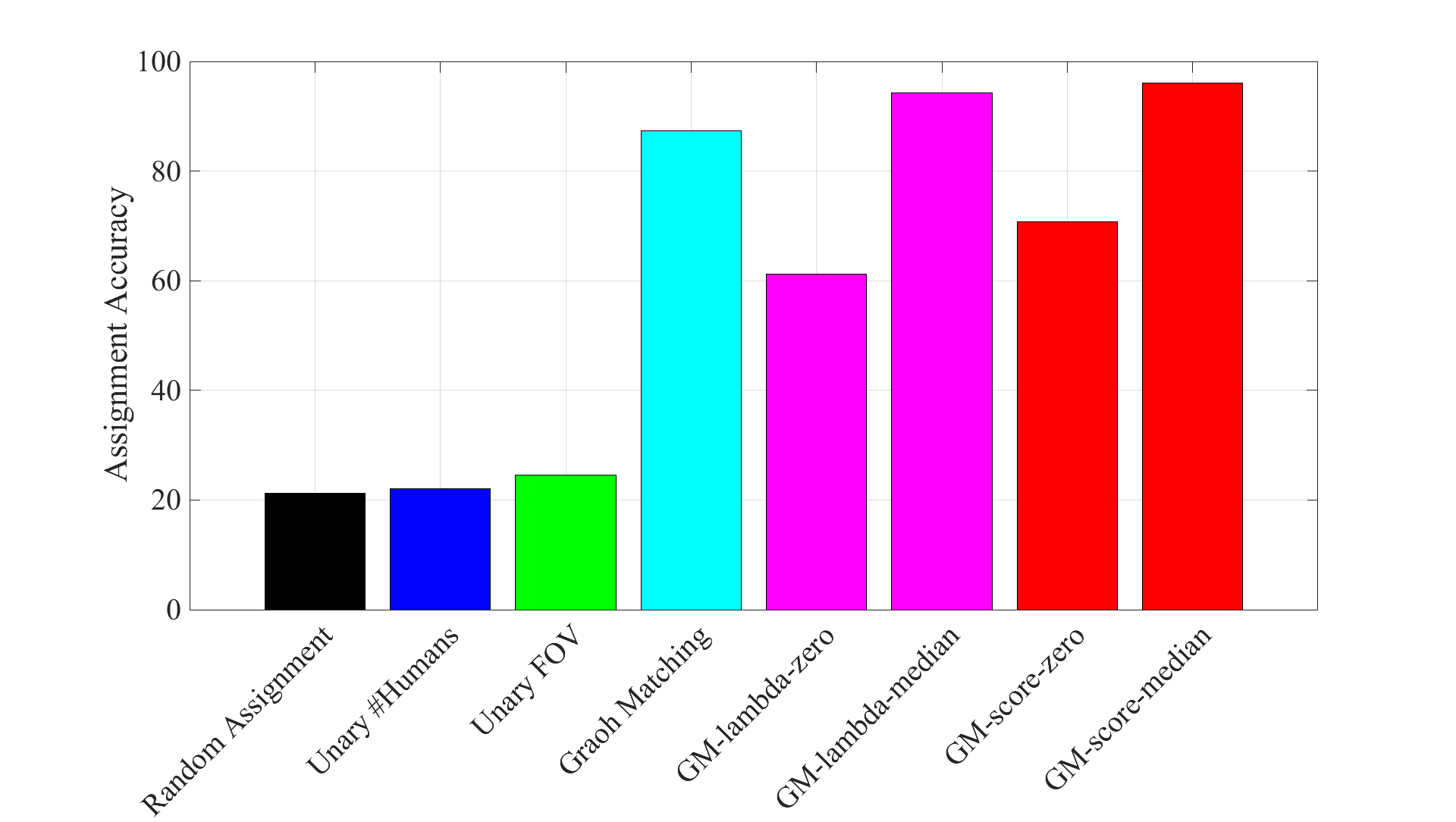}
			\caption{}
			\label{fig:fullSet_bar}			
		\end{subfigure}
		\caption{(a) shows the cumulative matching curve for ranking the viewers in the top-view video. The green and blue curves belong to ranking based on the cross correlation between the 2D, and cross correlation between the 1D unary scores, respectively (Not incorporating pairwise features). The cyan curve is the graph matching method results (section \ref{sec:ECCV_method}), and the magenta and red curves are the results of the two iterative approaches introduced section \ref{sec:PAMI_method} with two different initializations. The dashed black line shows random ranking accuracy (b) shows the assignment accuracy based on randomly assigning, using the number of humans, using unary features, using spectral graph matching, and using our two iterative approaches with two different initializations. The best performance in both (a) and (b) is achieved by the matching score based iterative optimization, when the time-delays are initialized by the median of the suggested values.}	
		\label{fig:all_results}			
	\end{center}
\end{figure*}

In order to answer the second question, we need to assess the accuracy of our method in terms of hard-assignment. Having a set of egocentric videos and a top-view video corresponding to the egocentric viewers, we compute the percentage of egocentric videos that were correctly matched to their corresponding viewer. We compare the hard-assignment accuracies of our two proposed algorithms with two different initializations, with four baselines in Figure \ref{fig:all_results}(b). Similar to the ranking performance, the first baseline is random assignment. For that purpose we randomly assign each egocentric video to one of the visible viewers in the top-view video. The second baseline is performing Hungarian bipartite matching only on the 1D unary feature which is the count of visible humans over times. The third baseline is performing Hungarian bipartite matching only on the 2D unary feature (GIST vs. FOV, denoted as Unary FOV), ignoring the pairwise relationships (edges) in the graphs. The fourth baseline is Graph Matching method introduced in \cite{ardeshir2016ego2top}. The consistent improvement of the Graph Matching method using both unary and pairwise features (denoted as GM) over the baselines shows the significant contribution of pairwise features in the assignment accuracy. The last four columns show the assignment accuracies using the two iterative algorithms proposed in this work. It shows that initializing the time delays as a vector of zeros would not improve the assignment accuracy. Instead using the median of suggested time-delays introduced in section \ref{sec:PAMI_method} will boost the assignment accuracy significantly. The highest accuracy is achieved by median-based initialization and the graph matching score using iterative-alternative algorithm, which results in $96\%$ assignment accuracy. The promising accuracy acquired by graph matching answers the second question. Knowing a top-view camera is capturing a set of egocentric viewers, we can use visual cues in the egocentric videos and the top-view video to decide reliably which viewer is capturing which egocentric video.


\subsubsection{Effect of Number of Egocentric Cameras:} In Sections 4.2.2 and 4.2.3, we evaluated the performance of our method given all the available egocentric videos present in each set as the input to our method. In this experiment, we compare the accuracy of our assignment and ranking framework as a function of the \textit{completeness ratio} ($\frac{n_{Ego}}{n_{Top}}$) of our egocentric set. Each of our sets contain $3<N^t<11$ viewers in the top-view camera, and $2<N^e<8$ egocentric videos. We evaluate the accuracy of our method and baselines using different subsets of the egocentric videos. 
A total of $2^{N^e}-1$ non-empty subsets of egocentric videos is possible 
depending on which egocentric video out of $N^e$ are included (all possible non-empty subsets). 

\begin{figure*}
	\begin{center}
		\begin{subfigure}[t]{0.24\textwidth}
			\includegraphics[width=\textwidth]{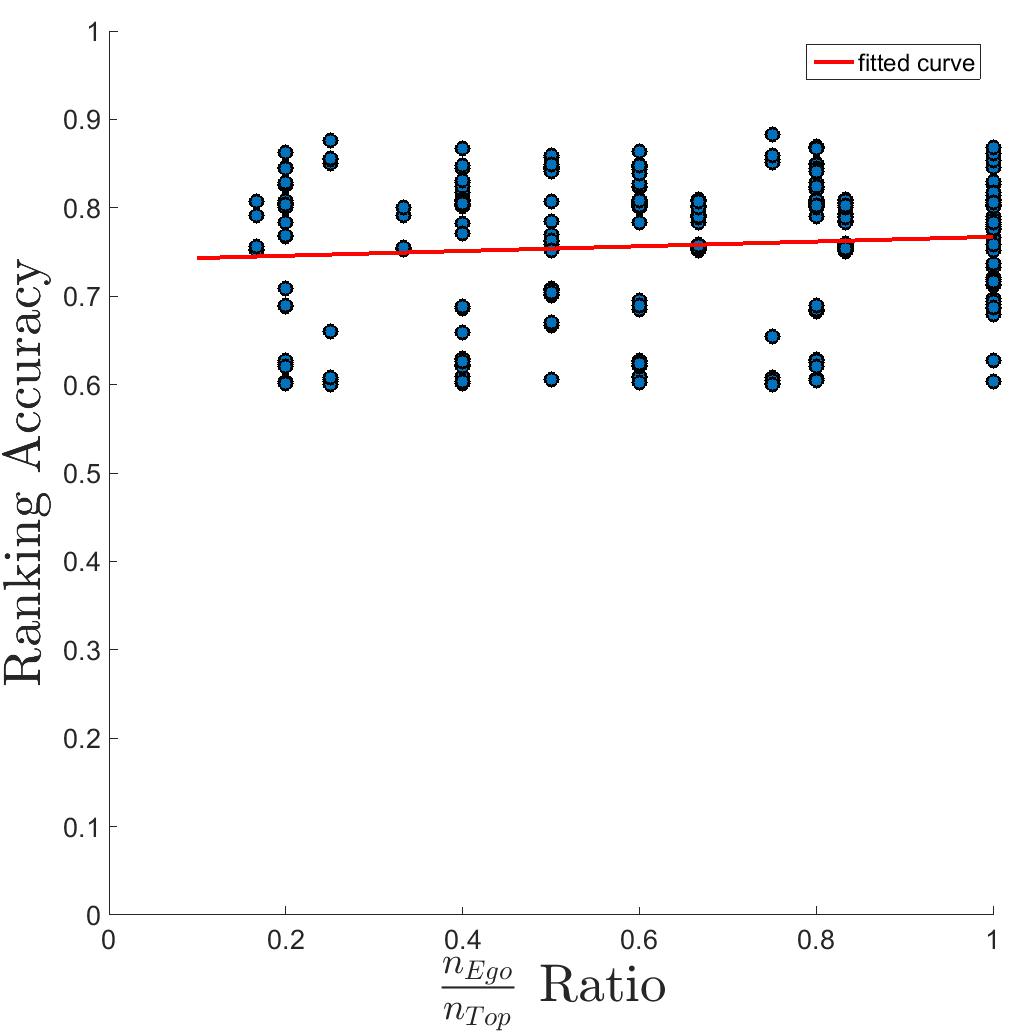}
			\caption{}			
			\label{fig:CMC_car_overall}			
		\end{subfigure}\hfill	
		\begin{subfigure}[t]{0.24\textwidth}
			\includegraphics[width=\textwidth]{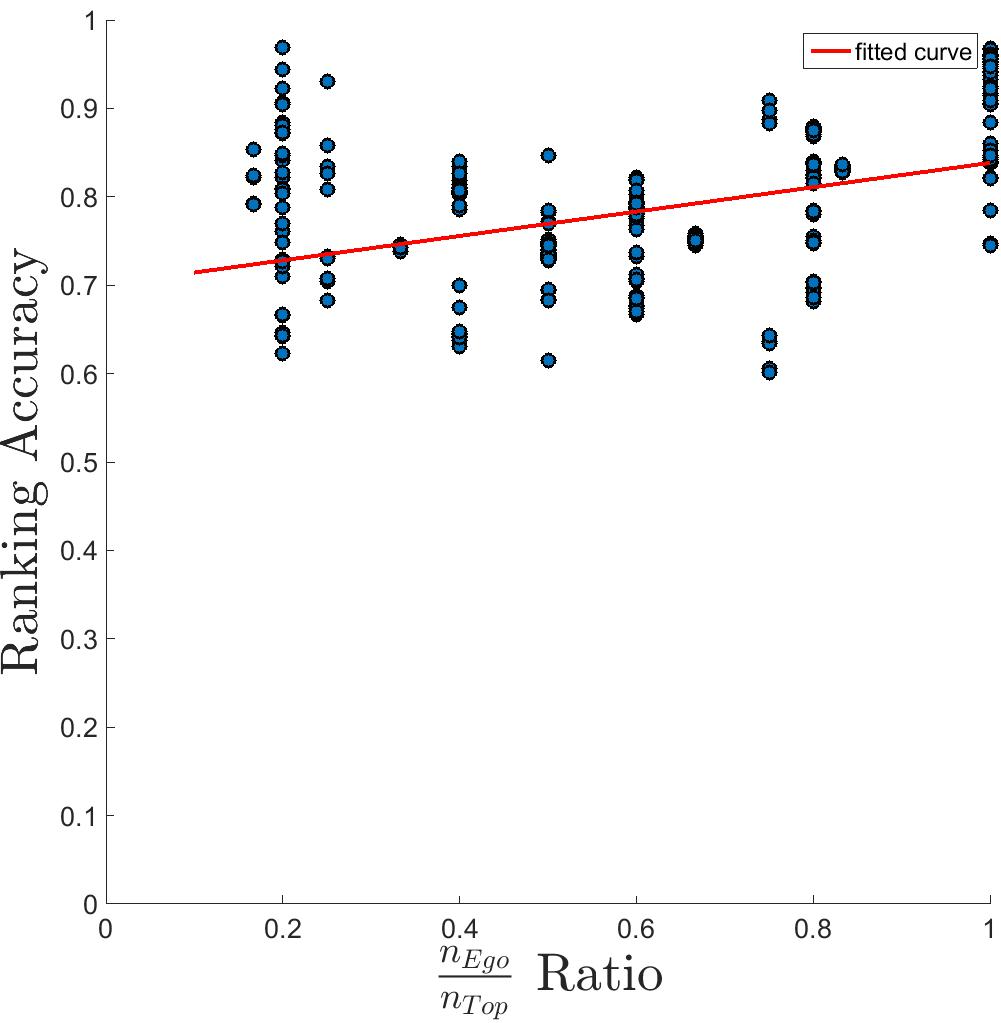}
			\caption{}
			\label{fig:CMC_car_overall}			
		\end{subfigure}\hfill
		\begin{subfigure}[t]{0.24\textwidth}
			\includegraphics[width=\textwidth]{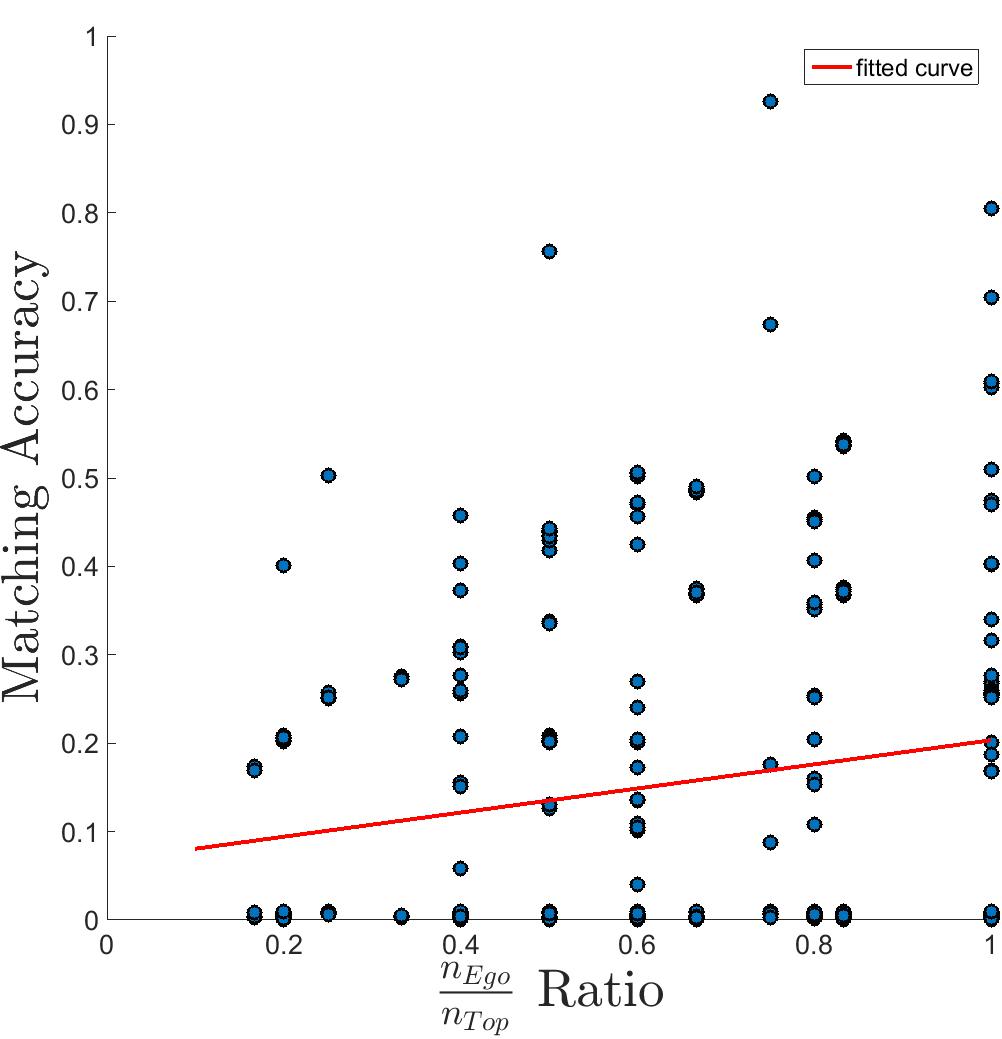}
			\caption{}		
			\label{fig:CMC_car_overall}			
		\end{subfigure}\hfill			
		\begin{subfigure}[t]{0.24\textwidth}
			\includegraphics[width=\textwidth]{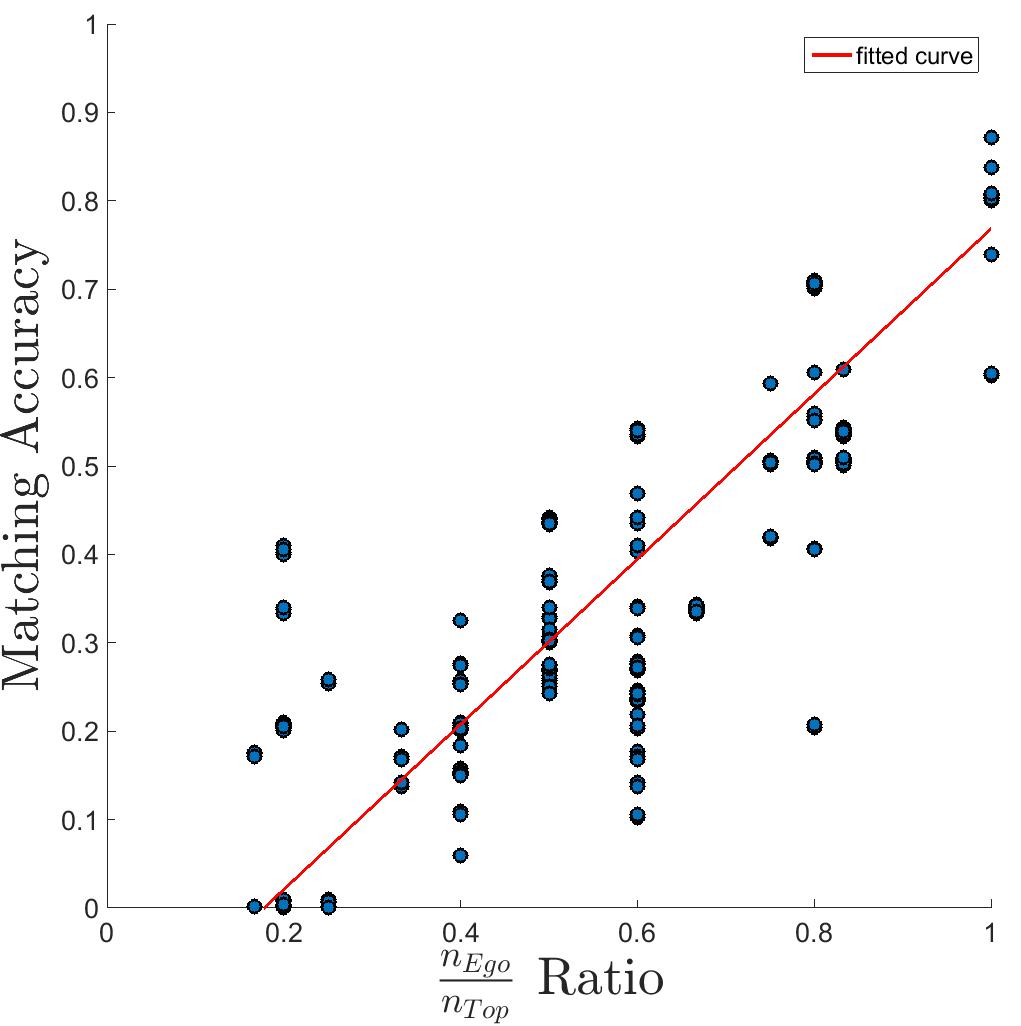}
			\caption{}			
			\label{fig:CMC_car_overall}			
		\end{subfigure}\hfill
		\caption{Effect of the relative number of egocentric cameras referred to as the completeness ratio ($\frac{n_{Ego}}{n_{Top}}$). (a) shows the ranking accuracy vs $\frac{n_{Ego}}{n_{Top}}$, only using the unary features, (b) shows the same evaluation using the graph matching output, (c) shows the accuracy of the hard assignment computed based on Hungarian bipartite matching on top of the unary features, and (d) shows the hard-assignment computed based on the spectral graph matching.}	
		\label{fig:Ego2TopRatio}			
	\end{center}
\end{figure*}

Figure \ref{fig:Ego2TopRatio} illustrates the assignment and ranking accuracies using the graph matching method \cite{ardeshir2016ego2top} versus the ratio of the available egocentric videos to the number of visible people in the top-view camera. It shows that as the completeness ratio increases, the assignment accuracy drastically improves. Intuitively, having more egocentric cameras gives more information about the structure of the graph (by providing more pairwise terms) which leads to improvement in the spectral graph matching and assignment accuracy. 

\begin{figure*}
	\begin{center}
			\begin{subfigure}[t]{0.49\textwidth}
			\includegraphics[width=\textwidth]{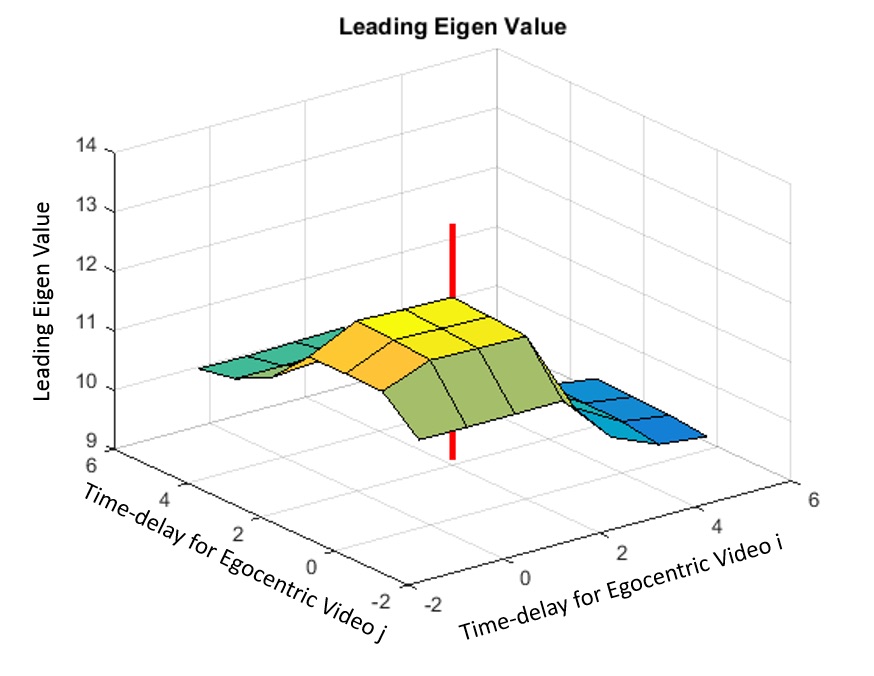}
			\caption{}
			\label{fig:CMC_car_overall}			
		\end{subfigure}\hfill
		\begin{subfigure}[t]{0.49\textwidth}
			\includegraphics[width=\textwidth]{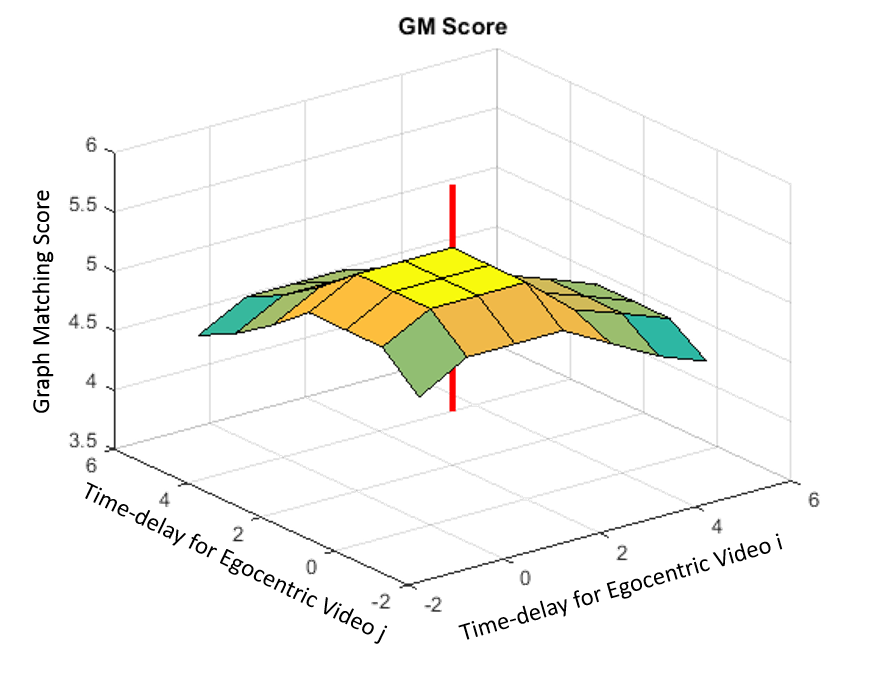}
			\caption{}		
			\label{fig:CMC_car_overall}			
		\end{subfigure}\hfill			
		\caption{An example of the objective function with respect to the time-delays of two egocentric videos. The z axis in both figures shows the objective to be maximized in each of the algorithms. The z axis in (a) shows the change in the leading eigenvalue of an affinity matrix with respect to the time delay of two of the egocentric videos. The red vertical line corresponds to the correct time delays, (b) shows the change in the graph matching score with respect to two time delays. It can be observed that both the leading eigenvalue and the graph matching score reach their maximum values if the correct time-delays are selected.}	
		\label{fig:timeDelayVsNoise}			
	\end{center}
\end{figure*}

\subsubsection{Effect of Video Length in Assignment Accuracy}
\begin{figure}
	\begin{center}
		\begin{subfigure}[t]{0.49\textwidth}
			\includegraphics[width=\textwidth]{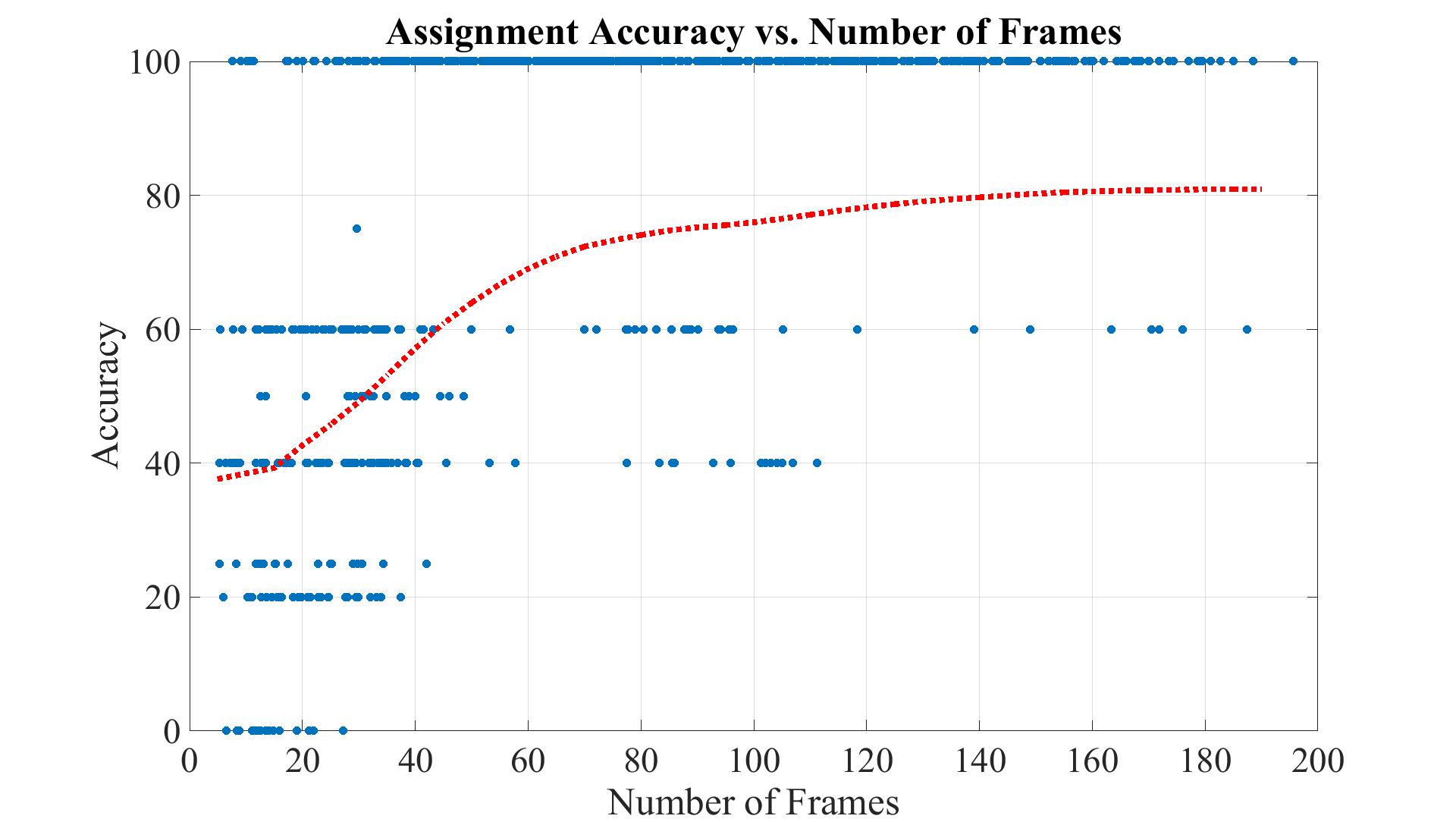}
		\end{subfigure}\hfill	
        \label{fig:timeAcc}	
		\caption{The effect of video length on the assignment accuracy. As the video length increases, the assignment accuracy improves. Please note that each blue dot represents one test sample, and the dashed red curve represents the cumulative mean assignment accuracy (mean of all accuracies with video length smaller than t)}	
		\label{fig:timeDelayVsNoise}			
	\end{center}
\end{figure}

Here, we analyze the effect of video length in assignment accuracy \cite{ardeshir2016ego2top}. For that purpose we use smaller portions of the videos and measure how the assignment accuracy changes as we use longer clips. As shown in Figure \ref{fig:timeAcc}, as the video length grows, the assignment accuracy increases. Intuitively, longer videos result in more discriminative unary and pairwise features and therefore lead to better performance.

\section{Conclusion and Discussion}
\label{sec:conclusion}
In this work, we addressed two main questions regarding relating multiple egocentric videos to a single top-view video. First, can we tell if a set of egocentric videos belong to a set of humans present in a top-view video? And second, given that they do, can we identify people? We proposed a unified framework that can properly answer these questions with high accuracy. 

Our experiments suggest that capturing the pattern of change in the content of the egocentric videos, along with capturing the relationships among them can help identify the viewers in top-view. To do so, we utilized a spectral graph matching technique and showed that the graph matching score is a meaningful criterion for narrowing down the search space in a set of top-view videos. Further, the assignment obtained by our framework is capable of associating egocentric videos to the viewers in the top-view camera. We conclude that meaningful features can be extracted from single, and pairs of egocentric camera(s), simply based on global scene gist of the content of the camera and incorporating the temporal information of the video(s).  

Empirical investigation shows that the assignment accuracy drops significantly if we do not include the binary features. This means that capturing the relationship among the viewers in top and egocentric views is an important factor. Also, enforcing consistency among the time-delays improved the accuracy in terms of assignment and ranking, as it prevents the system from producing invalid answers with contradictory implicit time-delay assignments. We demonstrate that the completeness of the egocentric set is a key factor in the performance of our proposed algorithms. Generally, the more complete the egocentric set, the higher assignment and ranking accuracy of the graph matching method. Video length is another significant factor. Longer videos result in more discriminative patterns in 1D and 2D feature descriptors, and thus a more accurate assignment. 

Our work helps relate two sources of information which so far have been studied in isolation and infer new insights about the visual world from different perspectives. We studied human identification but the same method can be used for understanding behavior of other entities such as animals or cars. For future, a more general case of this problem can be explored such as assigning multiple egocentric viewers to viewers in multiple top-view cameras. Also, other approaches can be explored for solving the introduced problem or slight variations of it (e.g., supervised methods for understanding the unary and pairwise relationships). Further, other computer vision techniques such as visual odometry can be explored for relating the two sources. We attempted to approach this problem using odometry at first, however, the results were not accurate perhaps due to a lot of jitter in egocentric videos. Nonetheless, this can be another potential direction for further research in the future.

	\bibliographystyle{IEEEtran}
	\bibliography{Thesis_bib}

%
%

%

\begin{IEEEbiographynophoto}{Shervin Ardeshir}
Received his B.Sc. degree in Electrical Engineering from Sharif University of Technology, Tehran, Iran 2012. And his M.Sc. at University of Central Florida in 2016. He is currently a PhD student at UCF's Center for Research in Computer Vision (CRCV). He has published multiple papers on topics such as egocentric vision, location aware image understanding, image geo-localization. 
\end{IEEEbiographynophoto}

\begin{IEEEbiographynophoto}{Ali Borji}
received the BS and MS degrees in
computer engineering from the Petroleum University
of Technology, Tehran, Iran, 2001 and
Shiraz University, Shiraz, Iran, 2004, respectively.
He received the PhD degree in cognitive
neurosciences from the Institute for Studies in
Fundamental Sciences (IPM) in Tehran, Iran,
2009. He is currently an assistant professor at
Center for Research in Computer Vision, University of Central Florida. His research interests include visual
attention, visual search, machine learning, robotics, neurosciences, and
biologically plausible vision models. He is a member of the IEEE.
\end{IEEEbiographynophoto}





\end{document}